\documentclass{article}

\usepackage{iclr2027_conference,times}
\usepackage{amsmath}
\usepackage{amssymb}
\usepackage{bm}
\usepackage{graphicx}
\usepackage{float}
\usepackage{wrapfig}
\usepackage{needspace}
\usepackage{capt-of}
\usepackage{stfloats}
\usepackage{placeins}
\usepackage{xcolor}
\usepackage{colortbl}
\usepackage{xspace}
\usepackage{booktabs}
\usepackage{multirow}
\usepackage{makecell}
\usepackage{array}
\usepackage{tabularx}
\usepackage{siunitx}
\usepackage{microtype}
\usepackage{url}
\usepackage{hyperref}
\usepackage[capitalise]{cleveref}

\graphicspath{{pics/}}
\newcommand{\methodtext}{MixPT}
\newcommand{\papertitle}{\methodtext: Point Transformer with\\Interleaved Window and Global Attention}
\newcommand{\methodname}{\methodtext\xspace}
\newcommand{\methodS}{\methodtext-S\xspace}
\newcommand{\methodSstar}{\methodtext-S*\xspace}
\newcommand{\methodB}{\methodtext-B\xspace}
\newcommand{\methodL}{\methodtext-L\xspace}

\newcommand{\ScanNet}{ScanNet\xspace}
\newcommand{\ScanNetTwoHundred}{ScanNet200\xspace}
\newcommand{\nuScenes}{nuScenes\xspace}
\newcommand{\Waymo}{Waymo\xspace}
\newcommand{\SemanticKITTI}{SemanticKITTI\xspace}
\newcommand{\SemKITTI}{Sem.KITTI\xspace}

\newcommand{\mIoU}{mIoU\xspace}
\newcommand{\mAcc}{mAcc\xspace}

\newcommand{\unitsep}{\;}
\newcommand{\meterval}[1]{\ensuremath{#1\unitsep\mathrm{m}}}

\newcommand{\ie}{\emph{i.e.},\xspace}

\crefname{section}{Sec.}{Secs.}
\Crefname{section}{Sec.}{Secs.}
\crefname{subsection}{Sec.}{Secs.}
\Crefname{subsection}{Sec.}{Secs.}
\crefname{subsubsection}{Sec.}{Secs.}
\Crefname{subsubsection}{Sec.}{Secs.}
\crefname{figure}{Fig.}{Figs.}
\Crefname{figure}{Fig.}{Figs.}
\crefname{table}{Table}{Tables}
\Crefname{table}{Table}{Tables}
\crefname{equation}{Eq.}{Eqs.}
\Crefname{equation}{Eq.}{Eqs.}
\creflabelformat{equation}{#2(#1)#3}

\newcommand{\feature}{\boldsymbol{x}}
\newcommand{\updatedfeature}{\widetilde{\boldsymbol{x}}}
\newcommand{\pooledfeature}{{}^{\mathrm{c}}\feature}
\newcommand{\updatedpooledfeature}{{}^{\mathrm{c}}\updatedfeature}
\newcommand{\poolednumtokens}{N_{\mathrm{g}}}
\newcommand{\numtokens}{N}
\newcommand{\windowsize}{N_{\mathrm{w}}}
\newcommand{\numchannels}{C}
\newcommand{\attention}{\mathcal{A}}
\newcommand{\mlp}{\operatorname{MLP}}
\newcommand{\pool}{\mathcal{P}}
\newcommand{\unpool}{\mathcal{U}}
\newcommand{\layernorm}{\operatorname{LN}}
\newcommand{\realnumbers}{\mathbb{R}}

\title{\papertitle}

\author{
Minkyun Seo\textsuperscript{1} \quad Hyungtae Lim\textsuperscript{2} \quad Jaesik Park\textsuperscript{1} \\
\makebox[0pt][l]{\raisebox{-\normalbaselineskip}[0pt][0pt]{\hypersetup{hidelinks}\texttt{\{\href{mailto:minkyunseo00@gmail.com}{minkyunseo00},\href{mailto:hyungtaelim.us@gmail.com}{hyungtaelim.us}\}@gmail.com} \quad \href{mailto:jaesik.park@snu.ac.kr}{\texttt{jaesik.park@snu.ac.kr}}}}%
\textsuperscript{1}Seoul National University \quad \textsuperscript{2}Massachusetts Institute of Technology
}

\newif\ifarxivmode
\newcommand{\arxivmode}{\arxivmodetrue\iclrfinalcopy}
\arxivmode

\begin{document}

\maketitle
\ifarxivmode
\lhead{Preprint}
\fi

\begin{figure}[H]
    \centering
    \includegraphics[width=\linewidth]{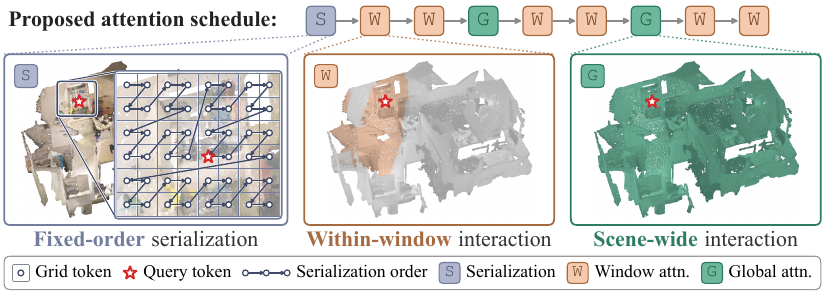}
   \caption{Serialize once, mix window and global attention.
\methodS{} serializes tokens once (\texttt{S}) and reuses that order throughout its \texttt{WWGWWGWW} attention schedule.
For the same query, window attention restricts interaction to its fixed window, whereas global attention enables scene-wide interaction.
For clarity, the left panel provides a 2D schematic of 3D serialization around the query token.}
\label{fig:teaser}
\end{figure}

\begin{abstract}

Point cloud transformers aim to capture detailed local geometry and relationships across a scene at a manageable computational cost.
Two common approaches are window attention and global attention.
They differ in how they enable scene-wide interaction, which refers to information exchange among tokens across a scene.
Window attention reduces computational cost by restricting self-attention to groups of tokens called windows; in serialized designs, each window contains a fixed number of consecutive tokens from a spatially ordered sequence.
Because each block only directly connects tokens within the same window, information exchange across windows relies on changes in window membership across blocks and hierarchical pooling.
Global attention, in contrast, directly connects all tokens within a scene, but computing attention over all token pairs can be costly when the number of tokens is large.
To balance computational cost and scene-wide interaction, we introduce \textit{\methodname}, a point transformer that uses a mix of window and global attention blocks in an interleaved schedule, while retaining a single fixed serialization pattern for its window blocks.
For scenes with a large number of tokens, \methodname applies global attention to spatially pooled tokens and returns residual updates to the original-resolution stream.
Across indoor and outdoor benchmarks, \methodname achieves competitive accuracy with lower latency than representative window and global attention baselines.
We further demonstrate the feasibility of scaling the attention stage in depth by repeating the interleaved pattern of window and global attention blocks.

\end{abstract}

\section{Introduction}
\label{sec:introduction}

Deep learning for point clouds has advanced through architectures that learn geometric representations from point sets or sparse spatial grids~\citep{Qi_2017_CVPR,NIPS2017_d8bf84be,Wang_2019,Wu_2019_CVPR,Thomas_2019_ICCV,Graham_2018_CVPR,Choy_2019_CVPR,Hu_2020_CVPR}.
More recently, point transformers~\citep{Zhao_2021_ICCV,NEURIPS2022_d78ece66,Wu_2024_CVPR} use attention to exchange information among tokens at different spatial locations.
At a fixed spatial resolution (\ie after voxel sampling with the same voxel size), larger 3D scenes typically retain more points and therefore require more input tokens, increasing the computational cost of attention.

Two common approaches make different choices in the trade-off between attention cost and scene-wide interaction: window attention and global attention.
Window attention reduces computational cost by restricting self-attention to groups of tokens called windows, so each block only directly connects tokens within the same window~\citep{Wu_2024_CVPR,Yue_2026_CVPR}.
Global attention, in contrast, directly connects all tokens within a scene, but computing attention over all token pairs can be costly when the number of tokens is large~\citep{yilmaz2026volt}.
This contrast motivates using a mix of window and global attention blocks: can window attention retain its computational efficiency while selected global blocks directly exchange information across windows?

Introducing global blocks also provides an opportunity to simplify how window blocks exchange information across windows.
Serialized window attention spatially orders tokens and groups consecutive tokens into fixed-size windows.
To indirectly exchange information across windows, these designs change serialization patterns across blocks to alter window membership and use hierarchical pooling to aggregate features at coarser resolutions~\citep{Wu_2024_CVPR,Yue_2026_CVPR}.
Keeping a single serialization pattern instead allows token ordering and window grouping to be reused across blocks, reducing processing overhead, but removes the interaction paths created by changing window membership.
Interleaving global attention blocks allows tokens in different windows to directly exchange information while keeping window membership fixed across window attention blocks.

We therefore introduce \methodname, a point transformer that uses a mix of window and global attention blocks in an interleaved schedule.
Its window blocks reuse a single fixed serialization pattern, while global blocks periodically enable direct information exchange across windows.
For scenes with a large number of tokens, \methodname applies global attention to spatially pooled tokens and returns only the resulting residual updates to the original-resolution stream.
This design retains window attention at the original stage resolution while controlling the cost of global attention through the frequency of global blocks and the number of pooled tokens.

\methodS{}, which is our smallest model variant, uses the \texttt{WWGWWGWW} attention schedule illustrated in \cref{fig:teaser}, where \texttt{W} and \texttt{G} denote window and global attention blocks, respectively.
Across indoor and outdoor 3D scene understanding benchmarks, \methodS{} achieves competitive accuracy with lower latency per forward pass than representative window and global attention baselines.
Our ablations show accuracy gains from periodically introducing global blocks compared with using fixed windows alone.
We further demonstrate the feasibility of increasing the attention stage's depth by repeating the interleaved pattern of window and global attention blocks.

In sum, our contributions are threefold.
First, we analyze how cross-window communication develops in serialized window attention and how the use of scene-wide support varies across global attention blocks.
Second, we propose \methodname, which interleaves fixed-order window and global attention, with pooled global attention for large token sets.
Finally, we demonstrate a favorable accuracy-latency trade-off across indoor and outdoor benchmarks and analyze the effects of block arrangement, pooling resolution, and model scaling.

\section{Related Work}
\label{sec:related-work}

\paragraph{Point Cloud Backbones.}
Early point-based backbones learned permutation-invariant features directly from unordered point sets and introduced hierarchical neighborhood aggregation to capture local geometry~\citep{Qi_2017_CVPR,NIPS2017_d8bf84be}.
Subsequent methods developed richer local point operators, scalable sampling strategies, and efficient model scaling~\citep{Wang_2019,Wu_2019_CVPR,Thomas_2019_ICCV,Hu_2020_CVPR,Thomas_2024_CVPR,NEURIPS2022_9318763d}, while sparse voxel approaches~\citep{Graham_2018_CVPR,Choy_2019_CVPR,Zhu_2021_CVPR} and an octree-based approach~\citep{Riegler_2017_CVPR} restricted feature storage and computation to occupied regions.
These approaches established local aggregation and hierarchical processing as effective foundations for large-scale point cloud understanding, which later point transformers build upon.

\paragraph{Point Cloud Transformers.}
Scaling point cloud transformers to large scenes requires controlling the number of point interactions and the cost of organizing them.
Early point transformers restrict attention to local neighborhoods, using nearest neighbors or spatial partitions to capture local geometry at manageable cost~\citep{Zhao_2021_ICCV,NEURIPS2022_d78ece66,Park_2022_CVPR}.
Later methods improve scalability by sharing spatial groups across queries~\citep{Fan_2022_CVPR,Lai_2022_CVPR,yang2023swin3dpretrainedtransformerbackbone}, adapting window geometry~\citep{Lai_2023_CVPR,NEURIPS2023_4b4f1272}, and forming fixed-size token groups for efficient batched attention~\citep{Liu_2023_CVPR,Wang_2023,Wang_2023_CVPR}.
PTv3~\citep{Wu_2024_CVPR} adopts serialized windows and changes serialization patterns across blocks to facilitate communication between groups.
We refer to its attention patches as \emph{windows} throughout this paper.
Efficiency also depends on the spatial resolution at which attention operates.
LitePT~\citep{Yue_2026_CVPR} uses sparse convolution at high resolution and retains serialized window attention in later stages.
Volt~\citep{yilmaz2026volt} instead applies global attention to volumetric patch tokens, controlling attention cost through token count rather than restricting the interaction range.

\paragraph{Interleaving Window and Global Attention.}
Interleaving local and global attention provides a way to preserve efficient local processing while introducing broader communication only at selected layers.
Global context can be introduced through compact representations that interact with local features~\citep{chu2021twins,ICLR2024_7e496423,pmlr-v202-hatamizadeh23a}, or through sparse connections between distant tokens~\citep{10.1007/978-3-031-20053-3_27}.
These designs reduce the cost of global communication while preserving detailed features within each window.
A complementary strategy retains full global attention but reserves it for selected layers, using window attention elsewhere, as adopted in vision backbones~\citep{10.1007/978-3-031-20077-9_17,Kirillov_2023_ICCV,ICLR2025_45c1f6a8,ICLR2026_e0982cbc,bai2025qwen25vltechnicalreport}.
This principle also extends to language modeling, where sliding window attention is interleaved with causal attention over the full context~\citep{gemmateam2024gemma2improvingopen,gemmateam2025gemma3technicalreport}.
However, the studies discussed above primarily explore this design principle in image and language models.
Thus, in this paper, we study its application to serialized point cloud backbones, focusing on fixed window grouping and the spatial resolution of global attention.

\paragraph{Rotary Positional Encoding for 3D Points.}
Rotary positional embedding (RoPE)~\citep{Su_2024} has been extended to 3D points using axial formulations~\citep{Li_2022_CVPR}.
Recent point cloud transformers use both symmetric~\citep{Yue_2026_CVPR,choy2026spaceformer,zhang2026utonia} and asymmetric~\citep{yilmaz2026volt} channel allocations across the three spatial axes.
We adopt an asymmetric allocation in \methodname and examine its effect in our ablations.

\section{Methodology}
\label{sec:method}

\subsection{Preliminaries: Revisiting Window and Global Attention}
\label{sec:preliminaries}
\begin{figure}[H]
    \centering
    \newlength{\reachimageheight}
    \setlength{\reachimageheight}{0.11771\linewidth}
    \newcommand{\reachpanel}[3]{%
        \begin{minipage}[t]{0.185\linewidth}
            \centering
            \parbox[c][\reachimageheight][c]{\linewidth}{%
                \centering
                #3}\par
            \vspace{2pt}
            {\fontsize{8}{10}\selectfont Attention block~#1: \texttt{#2}\par}
        \end{minipage}}
    \reachpanel{1}{Z}{\includegraphics[width=\linewidth]{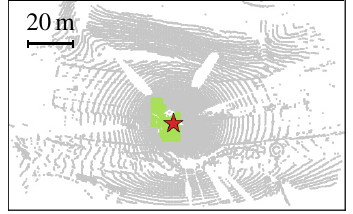}}\hfill
    \reachpanel{2}{TZ}{\includegraphics[width=\linewidth]{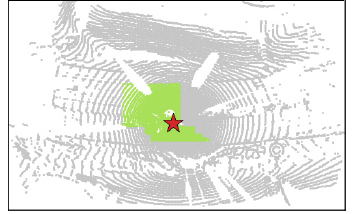}}\hfill
    \reachpanel{3}{H}{\includegraphics[width=\linewidth]{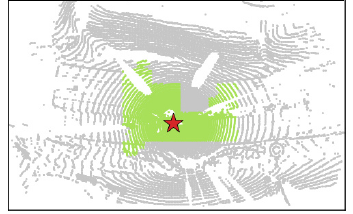}}\hfill
    \reachpanel{4}{TH}{\includegraphics[width=\linewidth]{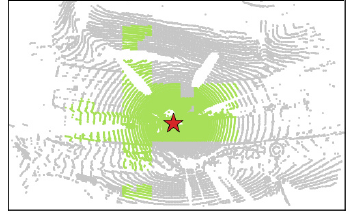}}\hfill
    \reachpanel{5}{Z}{\includegraphics[width=\linewidth]{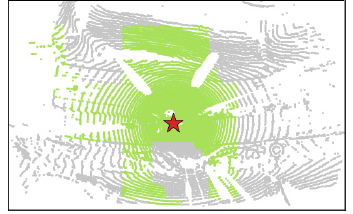}}
    \caption{Information propagation through the first five blocks of encoder stage~3 of LitePT on the \Waymo{} validation set~\citep{Sun_2020_CVPR}
(stages indexed from zero; blocks indexed from one; random order shuffling disabled).
Green marks tokens reachable from the red-star token through attention connections within this stage; gray marks those not yet reached.}
\label{fig:attention-reachability}
\end{figure}
\vspace{-0.3cm}

\paragraph{Revisiting Window Attention.}
Given \(\numtokens\) tokens, window attention partitions them into windows of \(\windowsize\) tokens each and computes self-attention only within each window, reducing the number of pairwise interactions from \(\mathcal{O}(\numtokens^2)\) to \(\mathcal{O}(\numtokens\windowsize)\).
Because tokens in different windows cannot interact directly, serialized window transformers rely on changes in serialization pattern and hierarchical pooling for cross-window communication~\citep{Wu_2024_CVPR,Yue_2026_CVPR}.
Alternating the serialization pattern changes window membership across blocks, while hierarchical pooling aggregates information at coarser resolutions.
PTv3~\citep[Table~2]{Wu_2024_CVPR} reports lower segmentation accuracy with a single fixed serialization pattern than with multiple patterns, suggesting that varying serialization patterns can be beneficial in practice.

\begin{wrapfigure}{r}{0.327273\columnwidth}
  \vspace{-3pt}
  \centering
  \setlength{\abovecaptionskip}{3pt}
  \includegraphics[width=\linewidth]{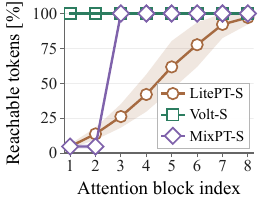}
  \caption{Structural reachability across blocks on the \Waymo{} validation set~\citep{Sun_2020_CVPR}.
Lines show scene-level means; shading indicates the 10th to 90th percentiles.}
  \label{fig:reachability-statistics}
\end{wrapfigure}

To assess these mechanisms, we examine how token reachability expands across successive blocks.
Here, a token is considered reachable if it can be connected to the selected token through preceding attention blocks and hierarchical pooling transitions.
We measure this structural reachability from the attention stage input tokens, excluding receptive fields accumulated in the preceding convolutional stages.
As shown in \cref{fig:attention-reachability} and \cref{fig:reachability-statistics}, LitePT-S~\citep{Yue_2026_CVPR}, a representative window attention backbone,
did not achieve 100\% mean token reachability on the \Waymo{} validation set~\citep{Sun_2020_CVPR} even in its deepest attention blocks, despite hierarchical pooling and changes in serialization pattern.
In contrast, Volt-S~\citep{yilmaz2026volt}, a representative global attention backbone, has full structural reachability from its first block.
These observations motivate introducing global attention early in the stage to establish full structural reachability without relying solely on gradual propagation through window attention and hierarchical transitions.
\methodS{} therefore places its first global block after two window blocks, while retaining a fixed serialization for its window blocks.

\WFclear
\paragraph{Revisiting Global Attention.}
\begin{wrapfigure}{r}{0.445455\columnwidth}
  \vspace{\dimexpr-\intextsep-4pt\relax}
  \centering
  \setlength{\abovecaptionskip}{3pt}
  \includegraphics[width=\linewidth]{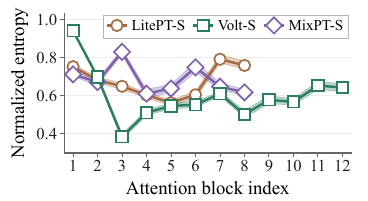}
  \caption{Normalized attention entropy within each block's available support on the \Waymo{} validation set~\citep{Sun_2020_CVPR}.
Lines show scene-level means; shading indicates the 10th to 90th percentiles.}
  \label{fig:attention-entropy}
\end{wrapfigure}

As shown in \cref{fig:reachability-statistics}, global attention achieves full structural reachability, but we ask whether its weights are also broadly distributed across tokens in every block.
To this end, we analyze a) how concentrated attention is within each block and b) where a selected query assigns its attention within the scene.
First, normalized attention entropy measures how evenly a query distributes attention over the keys it can attend to~\citep{Oh_2022_EMNLP,Zhai_2023_ICML}: a value of one indicates uniform attention over the available keys, while lower values indicate greater concentration on a subset.
For each block, we compute this quantity for every query and head, average over queries and heads within each scene, and then average the scene-level values equally across scenes.
\cref{fig:attention-entropy} shows higher entropy in the early Volt-S blocks and lower entropy in several later blocks, indicating that the concentration of attention varies across depth.

\WFclear
\begin{figure}[!b]
    \centering
    \newlength{\voltimageheight}
    \setlength{\voltimageheight}{0.11771\linewidth}
    \newcommand{\voltattentionmap}[2]{%
        \begin{minipage}[t]{0.185\linewidth}
            \centering
            \parbox[c][\voltimageheight][c]{\linewidth}{%
                \centering
                #2}\par
            \vspace{2pt}
            {\fontsize{8}{10}\selectfont Attention block~#1\par}
        \end{minipage}}
    \voltattentionmap{1}{\includegraphics[width=\linewidth]{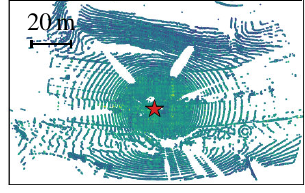}}\hfill
    \voltattentionmap{2}{\includegraphics[width=\linewidth]{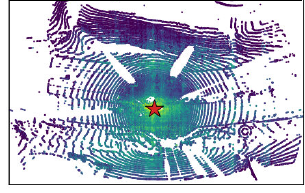}}\hfill
    \voltattentionmap{3}{\includegraphics[width=\linewidth]{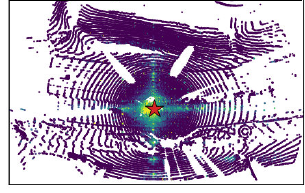}}\hfill
    \begin{minipage}[t]{0.02\linewidth}
        \centering
        \parbox[c][\voltimageheight][c]{\linewidth}{%
            \centering\makebox[\linewidth][c]{%
                \setlength{\unitlength}{1pt}%
                \hspace*{1.2pt}%
                \begin{picture}(7.5,1.4)
                    \put(0.85,0.7){\circle*{1.4}}%
                    \put(3.75,0.7){\circle*{1.4}}%
                    \put(6.65,0.7){\circle*{1.4}}%
                \end{picture}}}\par
        \vspace{2pt}
        {\fontsize{8}{10}\selectfont\strut\par}
    \end{minipage}\hfill
    \voltattentionmap{11}{\includegraphics[width=\linewidth]{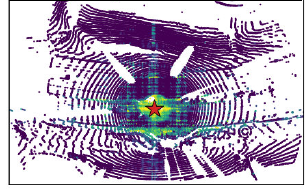}}\hfill
    \voltattentionmap{12}{\includegraphics[width=\linewidth]{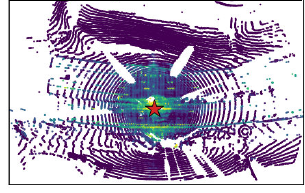}}\hfill
    \begin{minipage}[t]{0.045\linewidth}
        \centering
        \parbox[c][\voltimageheight][c]{\linewidth}{%
            \centering
            \includegraphics[height=\voltimageheight]{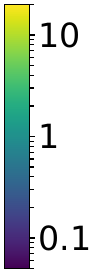}}\par
        \vspace{2pt}
        \makebox[\linewidth][c]{\scriptsize Rel.\ attn.}\par
    \end{minipage}
   \caption{Attention patterns across Volt~\citep{yilmaz2026volt} blocks on the \Waymo{} validation set~\citep{Sun_2020_CVPR}.
For a selected query, each token is colored by its relative attention weight (``Rel. attn.''), the ratio of its head-averaged attention weight to the uniform weight over the available keys.
Values of 1 and 10 indicate weights equal to and ten times the uniform weight, respectively.
Attention is broadly distributed in the first block and more concentrated in several later blocks, particularly in the central regions of the visualized scene.}
\label{fig:volt-attention}
\end{figure}

Second, we visualize the relative attention weights for a selected Volt query to show where attention concentrates within each block (\cref{fig:volt-attention}).
Although each block allowed global interactions, attention for this query tended to concentrate on localized regions in later blocks.
Together, these observations suggest that attention can become more concentrated at some depths despite having scene-wide support at every block.
These two key observations motivate the periodic use of global attention in \methodS{} for direct cross-window information exchange without computing attention between all token pairs in every block.

\subsection{\methodtext: Interleaved Window and Global Attention}

\begin{figure}[H]
    \centering
    \includegraphics[width=\linewidth]{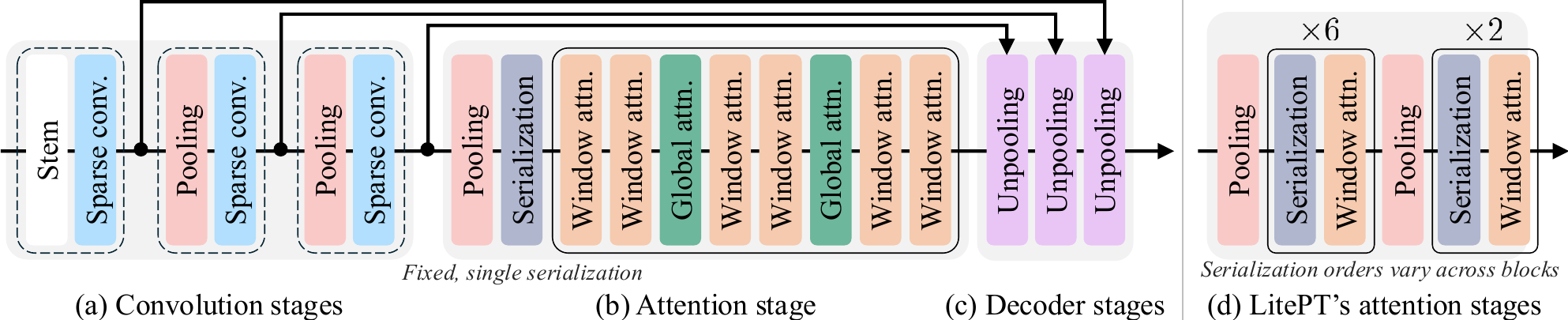}
    \caption{Block-level overview of \methodname: (a)~sparse convolution stages, (b)~attention stage, and (c)~decoder stages.
    Our attention stage follows the \texttt{WWGWWGWW} schedule, where \texttt{W} and \texttt{G} denote window and global attention blocks, respectively.
    (d) LitePT's attention stages~\citep{Yue_2026_CVPR}.
    Unlike LitePT's varying serialization patterns, \methodname uses one fixed serialization within a single attention stage and directly connects tokens across windows through global attention blocks.}
    \label{fig:method-overview}
\end{figure}

\paragraph{Architecture Overview.}
Panels (a)--(c) of \cref{fig:method-overview} illustrate the overall architecture of \methodname.
We retain the high-resolution sparse convolutional stages (\cref{fig:method-overview}(a)) and remaining decoder structure (\cref{fig:method-overview}(c)) of LitePT~\citep{Yue_2026_CVPR}.
Our main architectural change centers on the attention stage (\cref{fig:method-overview}(b)).
We consolidate LitePT's multiple attention stages (\cref{fig:method-overview}(d)) into a single stage by removing the intermediate pooling transition and the corresponding decoder upsampling.
We denote window and global attention blocks by \texttt{W} and \texttt{G}, respectively.
Within this stage, all window blocks reuse a single fixed serialization pattern, while global attention blocks are periodically interleaved following the \texttt{WWGWWGWW} schedule.
Both block types use the same number of heads and the same per-head dimension.
We next describe the fixed-order window and global attention blocks that form this stage.

\begin{wrapfigure}{r}{0.37\columnwidth}
    \vspace{-0.6\intextsep}
    \centering
    \includegraphics[width=0.49\linewidth]{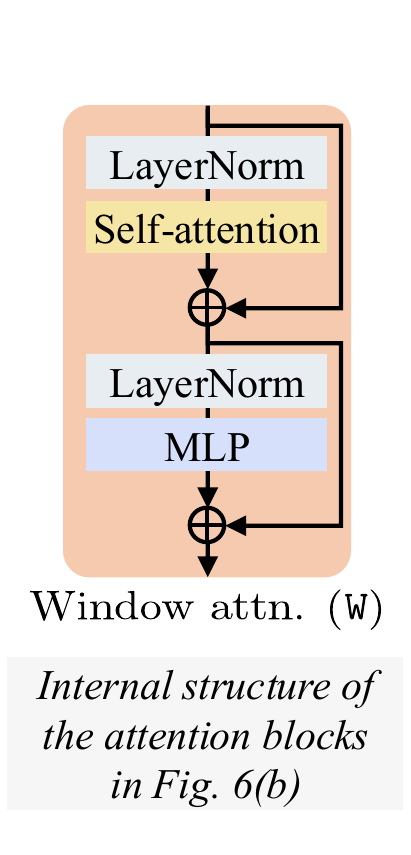}\hfill
    \includegraphics[width=0.49\linewidth]{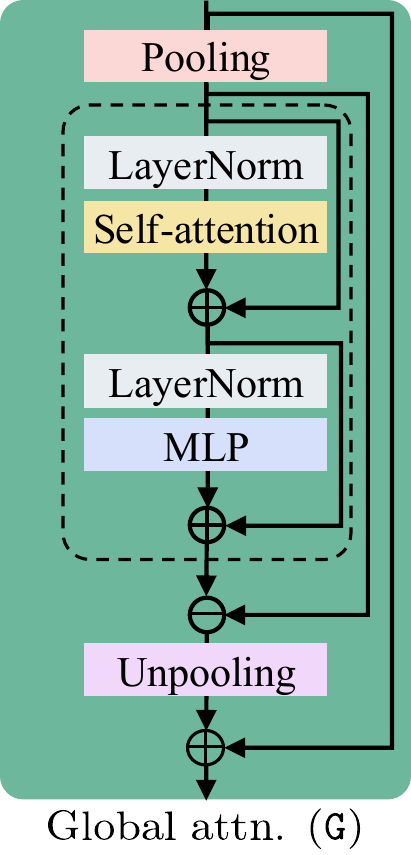}
    \setlength{\abovecaptionskip}{3pt}
    \caption{Window attention (left) and pooled global attention (right) blocks in \methodname.}
    \label{fig:attention-blocks}
\end{wrapfigure}

\paragraph{Fixed-Order Window Attention.}
We follow LitePT~\citep{Yue_2026_CVPR} in applying self-attention within serialized windows containing a fixed number of tokens, followed by an MLP.
We omit its additional block-entry LayerNorm on the residual stream and use a standard pre-norm design with LayerNorm only before the attention and MLP sublayers (\cref{fig:attention-blocks}).
Formally, let \(\feature_l \in \realnumbers^{\numtokens \times \numchannels}\) denote the input features to block~\(l\), where \(\numtokens\) is the number of tokens in the current attention stage for a scene and \(\numchannels\) is the feature dimension.
Let \(\windowsize\) denote the fixed number of tokens per window, and let \(\attention_l(\boldsymbol{X};m)\) denote self-attention applied independently within groups of \(m\) consecutive serialized tokens.
Let \(\mlp_l\) denote the MLP in block \(l\).
For window attention, we set \(m=\windowsize\) and apply the attention residual followed by the MLP residual:
\begin{align}
    \updatedfeature_l
    &= \feature_l
    + \attention_l\bigl(\layernorm(\feature_l);\windowsize\bigr),
    \label{eq:window-attention}\\
    \feature_{l+1}
    &= \feature_l
    + \attention_l\bigl(\layernorm(\feature_l);\windowsize\bigr)
    + \mlp_l\bigl(\layernorm(\updatedfeature_l)\bigr).
    \label{eq:window-block}
\end{align}
Unlike PTv3~\citep{Wu_2024_CVPR} and LitePT~\citep{Yue_2026_CVPR}, which vary serialization patterns across blocks, we use a single fixed pattern throughout the attention stage.
This allows the serialization and window-index mappings to be reused across window attention blocks.
Our default is Trans Z-order (\texttt{TZ}); we discuss sensitivity to this choice in Appendix~\ref{sec:suppl-serialization-order}.

\paragraph{Interleaved Global Attention with Pooling.}
A global block enables scene-wide communication across fixed windows.
When the number of stage tokens is manageable, we apply global attention directly at the original stage resolution.
Prior work has reduced the cost of long-range attention by representing distant or global context at a coarser spatial resolution~\citep{chu2021twins,yang2021focal,fan2021mvit}.
For large token sets, we apply global attention to spatially pooled tokens and return only the resulting residual updates to the original-resolution stream, as illustrated in \cref{fig:attention-blocks}.
We use \(s\) to denote the global pooling stride on the current attention-stage grid, with \(s=1\) indicating the unpooled case and \(s>1\) the pooled case.
Let \(\poolednumtokens\) denote the number of tokens used by global attention, with \(\poolednumtokens=\numtokens\) when \(s=1\).
For \(s>1\), \(\pool_s(\cdot)\) pools tokens within each stride-\(s\) grid cell, while \(\unpool_s(\cdot)\) maps coarse residual updates back to the corresponding original tokens.
These operators include the learned projections and pooling normalization omitted from the diagram.

Formally, the pooled global attention block is defined as
\par
\ifnum\value{WF@wrappedlines}>1
  \vspace{\dimexpr\value{WF@wrappedlines}\baselineskip-\baselineskip\relax}
\fi
\WFclear
\begin{align}
    \pooledfeature_l
    &= \pool_s(\feature_l), \qquad
    \updatedpooledfeature_l
    = \pooledfeature_l
    + \attention_l\bigl(
        \layernorm(\pooledfeature_l);
        \poolednumtokens
      \bigr),
    \label{eq:global-attention}\\
    \feature_{l+1}
    &= \feature_l
    + \unpool_s\!\Bigl(
        \attention_l\bigl(
            \layernorm(\pooledfeature_l);
            \poolednumtokens
        \bigr)
        + \mlp_l\bigl(
            \layernorm(\updatedpooledfeature_l)
        \bigr)
      \Bigr).
    \label{eq:global-block}
\end{align}
Here, the left superscript \(\mathrm{c}\) denotes features at the coarse resolution after pooling.
Only the attention and MLP residual updates are mapped back to the original tokens, preserving \(\feature_l\) as the outer residual stream.
We examine how the pooling stride affects segmentation accuracy and latency in Sec.~\ref{sec:ablation-studies}.

\paragraph{Rotary Positional Encoding for 3D Points.}
We follow the axial PointROPE formulation used in LitePT~\citep{Yue_2026_CVPR},
while using an asymmetric allocation of RoPE dimensions across the three spatial axes in our default configurations.
Specifically, one-dimensional RoPE is applied independently to the \(x\)-, \(y\)-, and \(z\)-axis subspaces of each attention head's query and key vectors.
Let \(d_h\) denote the per-head dimension, which is partitioned into axis-specific dimensions \(d_x\), \(d_y\), and \(d_z\), such that \(d_x+d_y+d_z=d_h\).
Allocations with \(d_x=d_y=d_z\) are considered symmetric; all others are asymmetric.
Each axis dimension is even to support the pairwise rotations in RoPE.
We examine the effect of the allocation in Sec.~\ref{sec:ablation-studies}.
\FloatBarrier

\section{Experiments}
\label{sec:experiments}

\subsection{Semantic Segmentation}

\begin{table*}[!htb]
  \caption{Median latency for a single model forward pass over voxelized validation scenes (ms), measured with batch size one and FP16 autocast on an RTX A6000.
  For \methodname, -S, -B, and -L denote the small, base, and large variants, respectively.}
  \label{tab:latency-comparison}
  \centering
  \small
  \setlength{\tabcolsep}{3.5pt}
  \begin{tabular}{lllrrrr}
    \toprule
    Method & Attention & \#Params & \SemKITTI{} & \Waymo & \nuScenes & \ScanNet \\
    \midrule
    PTv3~\citep{Wu_2024_CVPR} & Window & 46.1M & 63.6 & 87.8 & 38.3 & 48.1 \\
    LitePT-S~\citep{Yue_2026_CVPR} & Window & 12.7M & 36.4 & 53.0 & 27.8 & 25.7 \\
    Volt-S~\citep{yilmaz2026volt} & Global & 23.7M & 140.3 & 295.5 & 40.0 & 23.0 \\
    \textbf{\methodS{} (Ours)} & Window + Global & 14.1M & \textbf{28.2} & \textbf{42.8} & \textbf{19.2} & \textbf{17.5} \\
    \midrule
    \textbf{\methodB{} (Ours)} & Window + Global & 51.8M & 56.4 & 90.3 & 33.8 & 30.2 \\
    \textbf{\methodL{} (Ours)} & Window + Global & 89.4M & 72.4 & 120.1 & 45.7 & 33.4 \\
    \bottomrule
  \end{tabular}
\end{table*}

\begin{table*}[!htb]
  \setlength{\abovecaptionskip}{4pt}
  \caption{Semantic segmentation w/o TTA (\mIoU / \mAcc{}, \%).
  Bold and underline indicate the best and second-best results, respectively, among the baselines and \methodS{}.}
  \label{tab:semantic-segmentation}
  \centering
  \small
  \setlength{\tabcolsep}{4.1pt}
  \begin{tabular}{@{}lrrrrrrrrr@{}}
    \toprule
    & & \multicolumn{2}{c}{\SemKITTI{}} & \multicolumn{2}{c}{\Waymo} & \multicolumn{2}{c}{\nuScenes} & \multicolumn{2}{c}{\ScanNet} \\
    \cmidrule(lr){3-4}\cmidrule(lr){5-6}\cmidrule(lr){7-8}\cmidrule(l){9-10}
    Method & \#Params & \mIoU & \mAcc{} & \mIoU & \mAcc{} & \mIoU & \mAcc{} & \mIoU & \mAcc{} \\
    \midrule
    PTv3~\citep{Wu_2024_CVPR} & 46.1M & 69.6 & \underline{77.2} & 70.7 & 81.3 & 79.6 & 86.6 & \textbf{76.5} & \textbf{85.3} \\
    LitePT-S~\citep{Yue_2026_CVPR} & 12.7M & \underline{69.8} & 76.8 & \underline{71.7} & \underline{82.8} & 80.3 & 87.1 & 75.5 & 83.4 \\
    Volt-S~\citep{yilmaz2026volt} & 23.7M & 69.4 & 77.0 & 70.5 & 80.3 & \underline{80.5} & \textbf{87.5} & \underline{76.4} & \underline{84.9} \\
    \textbf{\methodS{} (Ours)} & 14.1M & \textbf{70.0} & \textbf{77.3} & \textbf{72.1} & \textbf{83.0} & \textbf{80.6} & \underline{87.4} & 76.0 & 84.3 \\
    \midrule
    \textbf{\methodB{} (Ours)} & 51.8M & 70.4 & 77.2 & 72.1 & 84.1 & 81.1 & 87.3 & 76.4 & 84.9 \\
    \textbf{\methodL{} (Ours)} & 89.4M & 70.6 & 77.4 & 72.4 & 83.7 & 81.5 & 87.9 & 76.7 & 85.1 \\
    \bottomrule
  \end{tabular}
\end{table*}

\paragraph{Setup.}
We evaluated semantic segmentation on the validation sets of \SemanticKITTI{}~\citep{Behley_2019_ICCV}, \Waymo{}~\citep{Sun_2020_CVPR}, \nuScenes{}~\citep{Caesar_2020_CVPR}, and \ScanNet{}~\citep{Dai_2017_CVPR}, covering both outdoor LiDAR scans and indoor RGB-D scenes.
We abbreviate \SemanticKITTI{} as \SemKITTI{} in all tables for brevity.
The comparison includes PTv3~\citep{Wu_2024_CVPR} and LitePT-S~\citep{Yue_2026_CVPR} as representative window attention baselines, and Volt-S~\citep{yilmaz2026volt} as a global attention baseline.
The main comparison reports results without test-time augmentation (TTA), using mean intersection over union (\mIoU{}) and mean class accuracy (\mAcc{}).
For semantic segmentation, we followed Volt~\citep{yilmaz2026volt} in using an exponential moving average~(EMA) of model weights and applied the same protocol to \methodname{} and all reproduced baselines.
Training, evaluation, and baseline reproduction details are provided in Appendix~\ref{sec:suppl-protocols}.
Full semantic segmentation results with TTA are reported in \cref{tab:semantic-segmentation-full}.

\begin{figure}[!htb]
  \centering
  \setlength{\abovecaptionskip}{3pt}
  \begin{minipage}[b]{0.49\linewidth}
    \centering
    \includegraphics[width=\linewidth]{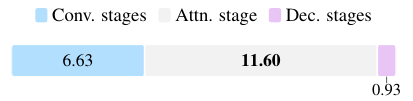}\par
    \vspace{1pt}
    {\normalsize\strut (a) Overall stage breakdown [ms]}
  \end{minipage}\hfill
  \begin{minipage}[b]{0.49\linewidth}
    \centering
    \includegraphics[width=\linewidth]{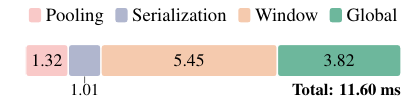}\par
    \vspace{1pt}
    {\normalsize\strut (b) Attention stage breakdown [ms]}
  \end{minipage}
  \caption{\methodS{} latency breakdown on \nuScenes{}: (a) across model stages and (b) across individual components within the attention stage.}
  \label{fig:latency-breakdown}
\end{figure}

\paragraph{Accuracy and Efficiency.}
First, we demonstrate how \methodname balances semantic segmentation accuracy and inference latency relative to window and global attention baselines.
As shown in \cref{tab:latency-comparison}, \methodS{} achieved the lowest latency across all four datasets, reducing latency by 19\% to 32\% relative to LitePT-S.
The gap was larger relative to Volt-S, the global attention baseline, on outdoor datasets where \methodS{} was 2.1 to 6.9$\times$ faster.
A key source of efficiency is that \methodname constructs the serialization order and stage layout once and reuses them across window attention blocks.
Token reorganization therefore occurs only at stage entry and exit, whereas LitePT repeatedly gathers and scatters tokens across blocks.
On \nuScenes{}, these serialization-related operations took 1.01\,ms in \methodS{}~(\cref{fig:latency-breakdown}), compared with 10.61\,ms in LitePT-S; further analysis is provided in Appendix~\ref{sec:suppl-serialization-order}.
Since the semantic segmentation head contributes only 0.04--0.25\% of the full forward latency, the measured latency primarily reflects backbone computation.
Despite these latency reductions, \methodS{} matched or improved upon LitePT-S in \mIoU{} across all four datasets when evaluated without TTA (\cref{tab:semantic-segmentation}).
Together, these results demonstrate a favorable accuracy-latency trade-off across both indoor and outdoor scenes.

\paragraph{Scaling the Model.}
We scaled \methodname by increasing model depth and per-head dimension while retaining 16 attention heads.
\methodB{} and \methodL{} generally improved performance across the evaluated benchmarks (\cref{tab:semantic-segmentation}).
Detailed model configurations are provided in \cref{tab:model-configuration} in Appendix~\ref{sec:suppl-architecture}.

\subsection{Instance Segmentation}

\begin{table*}[!htbp]
  \setlength{\abovecaptionskip}{4pt}
  \caption[Indoor instance segmentation on \ScanNet{} and \ScanNetTwoHundred{} (\%).]{\centering
  Indoor instance segmentation on \ScanNet{} and \ScanNetTwoHundred{} (\%).\par
  We evaluate different backbones using PointGroup~\citep{Jiang_2020_CVPR}.}
  \label{tab:instance-segmentation}
  \centering
  \small
  \setlength{\tabcolsep}{5.0pt}
  \begin{tabular}{@{}lrrrrrrr@{}}
    \toprule
    & & \multicolumn{3}{c}{\ScanNet} & \multicolumn{3}{c}{\ScanNetTwoHundred{}} \\
    \cmidrule(lr){3-5}\cmidrule(l){6-8}
    Backbone & \#Params & mAP$_{25}$ & mAP$_{50}$ & mAP & mAP$_{25}$ & mAP$_{50}$ & mAP \\
    \midrule
    PTv3~\citep{Wu_2024_CVPR} & 46.2M & 77.5 & 61.7 & 40.9 & \underline{40.1} & \textbf{33.2} & \textbf{23.1} \\
    LitePT-S*~\citep{Yue_2026_CVPR} & 16.0M & \underline{78.5} & \underline{64.9} & \textbf{41.7} & \textbf{40.3} & \underline{33.1} & 22.2 \\
    Volt-S~\citep{yilmaz2026volt} & 23.7M & 77.4 & 62.7 & 40.2 & 38.2 & 32.1 & 21.1 \\
    \textbf{\methodSstar{} (Ours)} & 18.4M & \textbf{78.9} & \textbf{65.1} & \underline{41.1} & 38.5 & 32.5 & \underline{22.4} \\
    \bottomrule
  \end{tabular}
\end{table*}

\paragraph{Setup and Results.}
We evaluated \methodname on the validation sets of \ScanNet{} and \ScanNetTwoHundred{} using a common PointGroup~\citep{Jiang_2020_CVPR} head (\cref{tab:instance-segmentation}) to focus the comparison on backbone representations.
For Volt~\citep{yilmaz2026volt}, we replaced its original SPFormer~\citep{Sun_2023} task decoder with PointGroup.
Following the instance segmentation setup of LitePT~\citep{Yue_2026_CVPR}, we used \methodSstar{}, which extends \methodS{} with additional attention and convolution blocks, without applying EMA during instance segmentation training.
Overall, \methodSstar{} remained competitive with the baselines across both datasets, suggesting that the interleaved window-global design extends beyond semantic segmentation.
Model configurations are detailed in Appendix~\ref{sec:suppl-architecture}, while training and evaluation settings are provided in Appendix~\ref{sec:suppl-instance-segmentation}.

\subsection{Ablation Studies and Analysis}
\label{sec:ablation-studies}

We provide empirical support for four design choices: i) fixed serialization with periodic global attention; ii) the arrangement of window and global attention blocks; iii) the global pooling stride; and iv) the allocation of RoPE dimensions across spatial axes.

\begin{table*}[!htbp]
  \setlength{\abovecaptionskip}{4pt}
  \caption[Fixed serialization and global attention ablation (w/o TTA, \mIoU / \mAcc{}, \%).]{\centering
  Fixed serialization and global attention ablation (w/o TTA, \mIoU / \mAcc{}, \%).\par
  Attention-stage depth and width were held fixed across variants.}
  \label{tab:fixed-serialization-global}
  \centering
  \small
  \setlength{\tabcolsep}{2.6pt}
  {\footnotesize
  \texttt{Z}: Z-order; \texttt{TZ}: Trans Z-order; \texttt{H}: Hilbert; \texttt{TH}: Trans Hilbert; \texttt{W}/\texttt{G}: window/global attention.
  }
  \par\smallskip
  \begin{tabular}{@{}llrrrrrrrr@{}}
    \toprule
     &  & \multicolumn{4}{c}{\SemKITTI{} (\(s=4\))} & \multicolumn{4}{c}{\ScanNet{} (\(s=1\))} \\
    \cmidrule(lr){3-6}\cmidrule(l){7-10}
    Serialization & Block layout & \#Params & Latency (ms) & \mIoU & \mAcc{} & \#Params & Latency (ms) & \mIoU & \mAcc{} \\
    \midrule
    \texttt{Z} + \texttt{TZ} + \texttt{H} + \texttt{TH} & \texttt{WWWWWWWW} & 13.6M & 32.99 & 68.5 & 75.6 & 13.6M & 21.28 & 74.8 & 84.0 \\
    \texttt{TZ} & \texttt{WWWWWWWW} & 13.6M & 27.88 & 68.6 & 76.3 & 13.6M & \textbf{17.16} & 73.7 & 83.5 \\
    \rowcolor{black!10}
    \texttt{TZ} & \texttt{WWGWWGWW} & 14.1M & 28.20 & \textbf{70.0} & \textbf{77.3} & 13.6M & 17.48 & \textbf{76.0} & \textbf{84.3} \\
    - & \texttt{GGGGGGGG} & 15.6M & \textbf{27.36} & 68.0 & 75.8 & 13.6M & 17.60 & 75.7 & 84.0 \\
    \bottomrule
  \end{tabular}
\end{table*}

\paragraph{Fixed Serialization and Global Attention.}
As presented in \cref{tab:fixed-serialization-global}, fixing the serialization pattern reduced latency on both datasets, while leaving \mIoU{} nearly unchanged on \SemanticKITTI{} and decreasing it on \ScanNet{}.
Changing two of the eight window blocks to global blocks raised \mIoU{} to 70.0\% and 76.0\%, respectively, with only a small increase in full-model latency.
On \SemanticKITTI{}, the global-only reference was slightly faster but less accurate than the interleaved design.
Unlike the interleaved design, this variant applied attention only to pooled tokens, without window attention at the original stage resolution.
Overall, periodic global attention improved the accuracy of fixed serialization while preserving its latency advantage over multi-pattern window attention.

\begin{table*}[!htbp]
  \setlength{\abovecaptionskip}{4pt}
  \caption{Block arrangement ablation for semantic segmentation (w/o TTA, \mIoU / \mAcc{}, \%).}
  \label{tab:block-arrangement}
  \centering
  \small
  \setlength{\tabcolsep}{2.7pt}
  \begin{tabular}{@{}lrrrrrrrrrr@{}}
    \toprule
     &  &  & \multicolumn{4}{c}{\nuScenes} & \multicolumn{4}{c}{\ScanNet} \\
    \cmidrule(lr){4-7}\cmidrule(l){8-11}
    Block layout & \#W & \#G & \#Params & Latency (ms) & \mIoU & \mAcc{} & \#Params & Latency (ms) & \mIoU & \mAcc{} \\
    \midrule
    \rowcolor{black!10}
    \texttt{WWGWWGWW} & 6 & 2 & 14.1M & \textbf{19.16} & 80.6 & 87.4 & 13.6M & 17.48 & \textbf{76.0} & \textbf{84.3} \\
    \texttt{WWWGWWWG} & 6 & 2 & 14.1M & 19.39 & 80.4 & 87.1 & 13.6M & 17.61 & 75.1 & 83.9 \\
    \midrule
    \texttt{WWWWGGGG} & 4 & 4 & 14.6M & 21.05 & 80.0 & 87.2 & 13.6M & 17.44 & 75.6 & 83.9 \\
    \texttt{WGWGWGWG} & 4 & 4 & 14.6M & 21.46 & 79.7 & 86.9 & 13.6M & \textbf{17.42} & 75.6 & \textbf{84.3} \\
    \midrule
    \texttt{WWWGWWWGWW} & 8 & 2 & 17.1M & 21.00 & \textbf{80.8} & \textbf{87.6} & 16.6M & 18.54 & 75.6 & 83.6 \\
    \bottomrule
  \end{tabular}
\end{table*}

\paragraph{How Should Window and Global Blocks Be Arranged?}
We varied the number and placement of window and global blocks to examine their effect on accuracy and latency (\cref{tab:block-arrangement}).
Among the eight-block schedules, \texttt{WWGWWGWW} provided a favorable accuracy-latency trade-off on \nuScenes{} and achieved the highest \mIoU{} on \ScanNet{}.
Increasing the number of global blocks did not consistently improve accuracy and incurred additional latency on \nuScenes{}, supporting the use of global attention only periodically.
A deeper ten-block variant slightly improved \nuScenes{} accuracy, but with more parameters and higher latency.

\begin{figure}[!htbp]
  \centering
  \setlength{\abovecaptionskip}{3pt}
  \begin{minipage}[t]{0.35\linewidth}
    \centering
    \includegraphics[width=\linewidth]{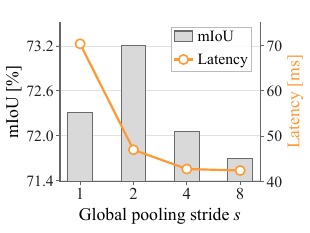}\par
    \vspace{1pt}
    {\normalsize\strut (a) Accuracy-latency trade-off}
  \end{minipage}\hfill
  \begin{minipage}[t]{0.305\linewidth}
    \centering
    \includegraphics[width=\linewidth]{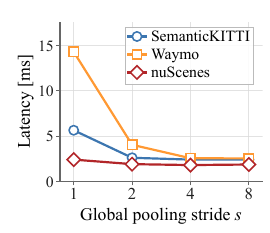}\par
    \vspace{1pt}
    {\normalsize\strut (b) Global block latency}
  \end{minipage}\hfill
  \begin{minipage}[t]{0.305\linewidth}
    \centering
    \includegraphics[width=\linewidth]{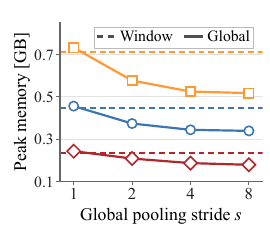}\par
    \vspace{1pt}
    {\normalsize\strut (c) Attn. block peak memory}
  \end{minipage}
  \caption{Effect of global pooling stride $s$ (w/o TTA).
  (a) \Waymo{} \mIoU{} (gray bars) and full-model latency (orange line), with separate training per stride under the same protocol.
  (b) Mean latency per global block.
  (c) Peak allocated memory for global (solid) and window (dashed) attention blocks.}
  \label{fig:global-pooling-stride}
\end{figure}

\paragraph{How Much Pooling Should Precede Global Attention?}
We examined how the global pooling stride \(s\) affected accuracy, latency, and memory.
On \Waymo{}, pooling reduced full-model latency and, at \(s=2\), even improved \mIoU{} over full-resolution global attention (\cref{fig:global-pooling-stride}(a)).
Global block latency decreased with pooling across all three datasets, with the largest absolute reduction on \Waymo{} (\cref{fig:global-pooling-stride}(b)).
For \(s \geq 2\), peak memory during global blocks fell below that of window blocks, while further pooling provided only modest additional reductions in global block memory (\cref{fig:global-pooling-stride}(c)).
Overall, these results suggest that global attention need not operate at full resolution.
An appropriate pooling stride can reduce its computational cost while preserving, or in some cases improving, segmentation accuracy.

\begin{table*}[!htbp]
  \setlength{\abovecaptionskip}{4pt}
  \caption{Axial RoPE allocation ablation (w/o TTA, \mIoU / \mAcc{}, \%).}
  \label{tab:rope-axis-allocation}
  \centering
  \small
  \setlength{\tabcolsep}{2.7pt}
  \begin{tabular}{@{}lrrrrrrrrrrr@{}}
    \toprule
     &  &  &  &  &  & \multicolumn{3}{c}{\nuScenes} & \multicolumn{3}{c}{\ScanNet} \\
    \cmidrule(lr){7-9}\cmidrule(l){10-12}
    RoPE & \#Heads & Head dim. & $d_x$ & $d_y$ & $d_z$ & \#Params & \mIoU & \mAcc{} & \#Params & \mIoU & \mAcc{} \\
    \midrule
    Axial (symmetric) & 16 & 18 & 6 & 6 & 6 & 10.0M & 79.8 & 87.3 & 9.7M & 75.2 & 84.0 \\
    Axial (symmetric) & 16 & 24 & 8 & 8 & 8 & 16.5M & 80.6 & 87.3 & 15.9M & 75.7 & 84.2 \\
    \midrule
    Axial (asymmetric) & 16 & 20 & 6 & 6 & 8 & 12.0M & 80.1 & 86.8 & 11.6M & 75.4 & \textbf{84.3} \\
    \rowcolor{black!10}
    Axial (asymmetric) & 16 & 22 & 8 & 8 & 6 & 14.1M & 80.6 & \textbf{87.4} & 13.6M & \textbf{76.0} & \textbf{84.3} \\
    Axial (asymmetric) & 16 & 26 & 10 & 10 & 6 & 19.1M & \textbf{80.8} & 87.2 & 18.4M & 75.5 & 84.0 \\
    \bottomrule
  \end{tabular}
\end{table*}

\paragraph{How Should RoPE Dimensions Be Allocated Across 3D Axes?}

\cref{tab:rope-axis-allocation} compares symmetric and asymmetric axial RoPE allocations and their model sizes.
The \(8{:}8{:}6\) allocation matched \(8{:}8{:}8\) on \nuScenes{} and outperformed it on \ScanNet{} with fewer parameters, consistent with the larger horizontal than vertical extents observed in the point clouds and attention regions.
\cref{fig:rope-window-geometry} shows representative examples, with validation set statistics provided in Appendix~\ref{sec:suppl-attention-geometry}.
These results suggest that RoPE dimensions need not be allocated symmetrically across the three spatial axes.
\begin{figure}[!htbp]
    \centering
    \setlength{\abovecaptionskip}{4pt}
    \includegraphics[width=\linewidth]{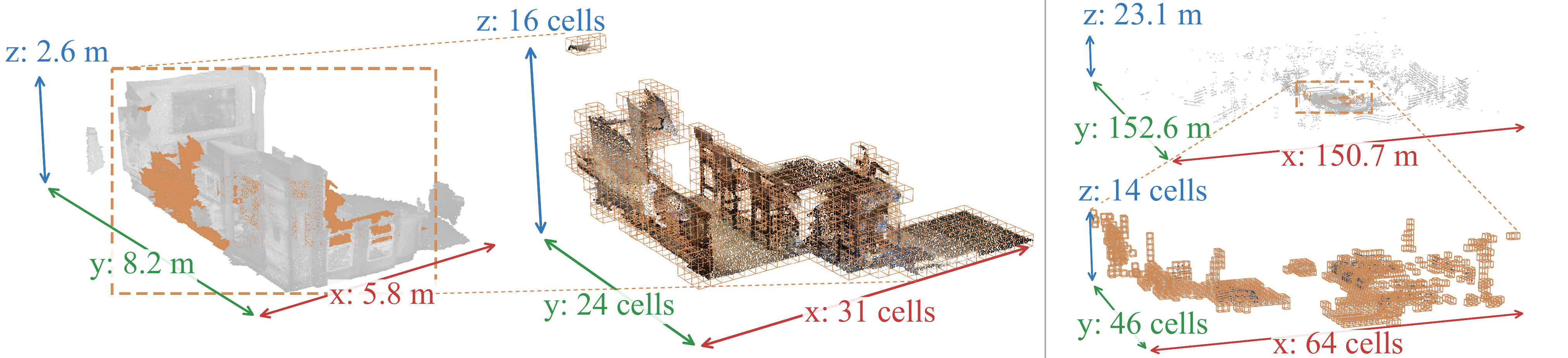}
    \caption{Spatial extents of representative scenes (metres) and serialized windows (grid cells) on \ScanNet{} (left) and \nuScenes{} (right).
    Dashed lines connect the selected window in each scene to its corresponding window view.
    Window extents include empty grid cells.}
    \label{fig:rope-window-geometry}
\end{figure}

\section{Conclusion}
\label{sec:conclusion}

We introduced \methodname, a point transformer that combines fixed-order window attention with periodically interleaved global attention blocks.
Our analysis motivates direct cross-window communication at selected layers, while spatial pooling reduces global attention cost for large token sets, with residual updates returned to the original-resolution stream.
Across indoor and outdoor 3D scene understanding benchmarks, \methodname achieves competitive accuracy, and \methodS{} shows lower per-forward latency than representative window and global attention baselines in semantic segmentation.
Ablations examine global block placement, pooling resolution, and scaling through repeated interleaved patterns, illustrating how \methodname controls when and at what spatial resolution global interaction occurs.

\clearpage

\bibliography{refs}
\bibliographystyle{iclr2027_conference}

\clearpage
\appendix
\setcounter{table}{0}
\renewcommand{\thetable}{A\arabic{table}}
\renewcommand{\theHtable}{supp.\arabic{table}}
\setcounter{figure}{0}
\renewcommand{\thefigure}{A\arabic{figure}}
\renewcommand{\theHfigure}{supp.\arabic{figure}}

\section{Architecture and Implementation Details}
\label{sec:suppl-architecture}

\begin{table*}[!htbp]
  \caption{\methodname{} model configurations.}
  \label{tab:model-configuration}
  \centering
  \small
  \setlength{\tabcolsep}{2pt}
  \renewcommand{\arraystretch}{1.15}
  \begin{tabularx}{\textwidth}{@{}l*{4}{>{\centering\arraybackslash}X}@{}}
    \toprule
    Setting & \methodS{} & \methodSstar{} & \methodB{} & \methodL{} \\
    \midrule
    Encoder channels (0--3) & $(36,72,144,352)$ & $(36,72,144,352)$ & $(54,108,216,512)$ & $(72,144,288,672)$ \\
    \specialrule{0.3pt}{0.8pt}{0.8pt}
    Encoder pooling strides & \multicolumn{4}{c}{$(2, 2, 2)$} \\
    \specialrule{0.3pt}{0.8pt}{0.8pt}
    Encoder blocks (0--3) & $(2, 2, 2, 8)$ & $(2, 2, 2, 10)$ & $(3, 3, 3, 14)$ & $(3, 3, 3, 14)$ \\
    \specialrule{0.3pt}{0.8pt}{0.8pt}
    Encoder attention & $(\mathtt{WWG})_2\mathtt{WW}$ & $(\mathtt{WWG})_3\mathtt{G}$ & $(\mathtt{WWG})_4\mathtt{WW}$ & $(\mathtt{WWG})_4\mathtt{WW}$ \\
    \specialrule{0.3pt}{0.8pt}{0.8pt}
    Decoder channels (2--0) & $(144,72,72)$ & $(144,72,72)$ & $(216,108,108)$ & $(288,144,144)$ \\
    \specialrule{0.3pt}{0.8pt}{0.8pt}
    Decoder conv. blocks (2--0) & $(0,0,0)$ & $(2,2,2)$ & $(0,0,0)$ & $(0,0,0)$ \\
    \specialrule{0.3pt}{0.8pt}{0.8pt}
    Window size ($\windowsize$) & \multicolumn{4}{c}{1{,}024} \\
    \specialrule{0.3pt}{0.8pt}{0.8pt}
    Attention heads & \multicolumn{4}{c}{16} \\
    \specialrule{0.3pt}{0.8pt}{0.8pt}
    Head dimension & 22 & 22 & 32 & 42 \\
    \specialrule{0.3pt}{0.8pt}{0.8pt}
    MLP expansion ratio & \multicolumn{4}{c}{4} \\
    \specialrule{0.3pt}{0.8pt}{0.8pt}
    RoPE $(d_x,d_y,d_z)$ & $(8,8,6)$ & $(8,8,6)$ & $(12,12,8)$ & $(16,16,10)$ \\
    \bottomrule
  \end{tabularx}
\end{table*}

\FloatBarrier

The notation \((\mathtt{WWG})_2\mathtt{WW}\), for example, expands to \(\mathtt{WWGWWGWW}\).
\cref{tab:model-configuration} lists the configurations for the architecture in \cref{fig:method-overview}; shared convolutional and decoder operators follow LitePT~\citep{Yue_2026_CVPR}.
\methodB{} and \methodL{} increase stage widths and convolutional depths and extend the attention schedule to \((\mathtt{WWG})_4\mathtt{WW}\).
For instance segmentation, \methodSstar{} extends the \methodS{} encoder attention stage from eight to ten blocks by appending two global blocks and adds two convolutional blocks after unpooling at each decoder stage.

For \(s>1\), global pooling applies a linear projection, channel-wise max pooling within each grid cell, and batch normalization.
A second linear projection maps the global block's residual update to the stage width; the update is then broadcast to the cell's member tokens and added to their features using the stored pooling assignments.
Dataset-specific pooling strides are listed in \cref{tab:outdoor-training-settings,tab:indoor-training-settings}.

\section{Experimental Protocols}
\label{sec:suppl-protocols}

\subsection{Semantic Segmentation}
\label{sec:suppl-semantic-segmentation}

\paragraph{Datasets and Training Settings.}
We evaluate on \SemanticKITTI{}~\citep{Behley_2019_ICCV}, \Waymo{}~\citep{Sun_2020_CVPR}, \nuScenes{}~\citep{Caesar_2020_CVPR}, and \ScanNet{}~\citep{Dai_2017_CVPR} under full supervision, following their standard training and validation splits.
We follow the training recipe of LitePT~\citep{Yue_2026_CVPR}, with dataset-specific settings summarized in \cref{tab:outdoor-training-settings,tab:indoor-training-settings}.
Following Volt~\citep{yilmaz2026volt}, we apply an exponential moving average (EMA) to model weights with a decay rate of \(0.999\) when training \methodname and reproducing all baselines.
All \methodname results reported in this section use the EMA weights.
For each baseline and dataset, we report the result with higher \mIoU{} between the original authors' report and our EMA-based reproduction, retaining the corresponding \mAcc{} from the same source.
We use cross-entropy with the Lov\'asz-Softmax loss~\citep{Berman_2018_CVPR}.
On \SemanticKITTI{}, we additionally use class weighting and InstanceCutMix~\citep{Xu_2021_ICCV}, applying the same protocol to \methodname and all reproduced baselines.

\begin{table}[!htb]
  \caption{Outdoor semantic segmentation training and evaluation settings for \methodS{}.}
  \label{tab:outdoor-training-settings}
  \centering
  \small
  \setlength{\tabcolsep}{5pt}
  \renewcommand{\arraystretch}{1.15}
  \begin{tabularx}{\linewidth}{@{}l*{3}{>{\centering\arraybackslash}X}@{}}
    \toprule
    Setting & \SemKITTI{} & \Waymo{} & \nuScenes{} \\
    \midrule
    Input feature & \multicolumn{3}{c}{XYZ + intensity} \\
    \specialrule{0.3pt}{0.8pt}{0.8pt}
    Grid size & \multicolumn{3}{c}{\meterval{0.05}} \\
    \specialrule{0.3pt}{0.8pt}{0.8pt}
    Global pooling stride ($s$) & 4 & 4 & 2 \\
    \specialrule{0.3pt}{0.8pt}{0.8pt}
    Head (framework) & \multicolumn{3}{c}{Linear segmentor} \\
    \specialrule{0.3pt}{0.8pt}{0.8pt}
    Loss & \makecell[tc]{Class-weighted\\cross-entropy\\+ Lov\'asz} & \makecell[tc]{Cross-entropy\\+ Lov\'asz} & \makecell[tc]{Cross-entropy\\+ Lov\'asz} \\
    \specialrule{0.3pt}{0.8pt}{0.8pt}
    Optimizer & \multicolumn{3}{c}{AdamW} \\
    \specialrule{0.3pt}{0.8pt}{0.8pt}
    Weight decay & \multicolumn{3}{c}{0.005} \\
    \specialrule{0.3pt}{0.8pt}{0.8pt}
    Scheduler & \multicolumn{3}{c}{OneCycleLR} \\
    \specialrule{0.3pt}{0.8pt}{0.8pt}
    Peak learning rate & \multicolumn{3}{c}{0.002} \\
    \specialrule{0.3pt}{0.8pt}{0.8pt}
    Block learning rate & \multicolumn{3}{c}{0.0002} \\
    \specialrule{0.3pt}{0.8pt}{0.8pt}
    Total batch size & 12 & 8 & 12 \\
    \specialrule{0.3pt}{0.8pt}{0.8pt}
    Epochs & 40 & 50 & 75 \\
    \specialrule{0.3pt}{0.8pt}{0.8pt}
    Number of GPUs & \multicolumn{3}{c}{2} \\
    \specialrule{0.3pt}{0.8pt}{0.8pt}
    EMA decay & \multicolumn{3}{c}{0.999} \\
    \specialrule{0.3pt}{0.8pt}{0.8pt}
    Data augmentation & \multicolumn{3}{c}{Random rotation, scaling, flips, jitter, scene mixing} \\
    & + InstanceCutMix & & \\
    \midrule
    TTA transformations & \makecell[tc]{Scales 0.90--1.10\\(step 0.05), each\\with / without XY flip} & \makecell[tc]{$z$-rotations:\\$0^\circ,90^\circ,180^\circ,270^\circ$} & \makecell[tc]{Scales 0.90--1.10\\(step 0.05), each\\with / without XY flip} \\
    \specialrule{0.3pt}{0.8pt}{0.8pt}
    TTA views & 10 & 4 & 10 \\
    \bottomrule
  \end{tabularx}
\end{table}

\begin{table}[!htb]
  \caption{Indoor training and evaluation settings for \methodS{} (semantic segmentation) and \methodSstar{} (instance segmentation).}
  \label{tab:indoor-training-settings}
  \centering
  \small
  \setlength{\tabcolsep}{5pt}
  \renewcommand{\arraystretch}{1.15}
  \begin{tabularx}{\linewidth}{@{}l*{3}{>{\centering\arraybackslash}X}@{}}
    \toprule
    \multirow{2}{*}{Setting} & \makecell{Semantic\\segmentation} & \multicolumn{2}{c}{Instance segmentation} \\
    \cmidrule(lr){2-2}\cmidrule(l){3-4}
    & \ScanNet{} & \ScanNet{} & \ScanNetTwoHundred{} \\
    \midrule
    Input feature & \multicolumn{3}{c}{RGB + normals} \\
    \specialrule{0.3pt}{0.8pt}{0.8pt}
    Grid size & \multicolumn{3}{c}{\meterval{0.02}} \\
    \specialrule{0.3pt}{0.8pt}{0.8pt}
    Global pooling stride ($s$) & \multicolumn{3}{c}{1} \\
    \specialrule{0.3pt}{0.8pt}{0.8pt}
    Head (framework) & Linear segmentor & PointGroup~\citep{Jiang_2020_CVPR} & PointGroup~\citep{Jiang_2020_CVPR} \\
    \specialrule{0.3pt}{0.8pt}{0.8pt}
    Loss & \makecell[tc]{Cross-entropy\\+ Lov\'asz} & \makecell[tc]{Cross-entropy\\+ Lov\'asz\\+ offset regression\\($L_1$ + cosine)} & \makecell[tc]{Cross-entropy\\+ Lov\'asz\\+ offset regression\\($L_1$ + cosine)} \\
    \specialrule{0.3pt}{0.8pt}{0.8pt}
    Optimizer & \multicolumn{3}{c}{AdamW} \\
    \specialrule{0.3pt}{0.8pt}{0.8pt}
    Weight decay & \multicolumn{3}{c}{0.05} \\
    \specialrule{0.3pt}{0.8pt}{0.8pt}
    Scheduler & \multicolumn{3}{c}{OneCycleLR} \\
    \specialrule{0.3pt}{0.8pt}{0.8pt}
    Peak learning rate & 0.006 & 0.006 & 0.01 \\
    \specialrule{0.3pt}{0.8pt}{0.8pt}
    Block learning rate & 0.0006 & 0.0006 & 0.001 \\
    \specialrule{0.3pt}{0.8pt}{0.8pt}
    Total batch size & \multicolumn{3}{c}{12} \\
    \specialrule{0.3pt}{0.8pt}{0.8pt}
    Epochs & 1200 & 800 & 800 \\
    \specialrule{0.3pt}{0.8pt}{0.8pt}
    Number of GPUs & \multicolumn{3}{c}{2} \\
    \specialrule{0.3pt}{0.8pt}{0.8pt}
    EMA decay & 0.999 & -- & -- \\
    \specialrule{0.3pt}{0.8pt}{0.8pt}
    Data augmentation & \multicolumn{3}{c}{\makecell[tc]{Center shift, dropout, random rotation, scaling, flips, jitter,\\elastic distortion, color auto-contrast, color translation/jitter,\\sphere crop, color normalization, scene mixing}} \\
    \midrule
    TTA transformations & \multicolumn{3}{c}{\makecell[tc]{Semantic segmentation only: $z$-rotations $0^\circ,90^\circ,180^\circ,270^\circ$\\at scales 0.95, 1.00, 1.05;\\one extra XY flip (scale 1, no rotation)}} \\
    \specialrule{0.3pt}{0.8pt}{0.8pt}
    TTA views & 13 & -- & -- \\
    \bottomrule
  \end{tabularx}
\end{table}

\paragraph{Evaluation.}
We follow the evaluation protocols of PTv3~\citep{Wu_2024_CVPR} and LitePT~\citep{Yue_2026_CVPR}, including their fragment-based prediction aggregation procedures.
A TTA view is a transformed version of the input scene, whereas a fragment is a voxelized realization of a view obtained by selecting different representative points from occupied voxels; fragments do not spatially partition the scene.
In addition to results with TTA, we report results without TTA by disabling all test-time transformations and using only the identity view.
Voxelization, fragment construction, and prediction aggregation remain unchanged.
The dataset-specific TTA settings follow these protocols and are summarized in \cref{tab:outdoor-training-settings,tab:indoor-training-settings}.
The listed view counts include the identity view; XY flip denotes negating both horizontal coordinates.
For the outdoor datasets, an initial 0.025\,m grid sampling reduces the input while retaining a mapping to the original points.
Class probabilities are summed at corresponding point indices across fragments and views, and the resulting predictions are mapped back to the original validation points to compute \mIoU{} and \mAcc{}.
Thus, evaluation without TTA can still require multiple forward passes, as illustrated in Appendix~\ref{sec:suppl-tta-fragments}.

\subsection{Efficiency Measurement}
\label{sec:suppl-efficiency}
Latency in \cref{tab:latency-comparison} is measured on an NVIDIA RTX A6000 with batch size one and FP16 autocast, using evaluation mode with gradients disabled.
All methods use the native sparse convolution kernels in spconv.
We time one complete model forward per voxelized validation scene, excluding input preprocessing, data transfer, loss computation, and prediction aggregation.
After warm-up on eight scenes, each input is timed three times using synchronized CUDA events, and we report the median over the validation split of the per-scene median latency.
These timings exclude repeated forwards for fragments and TTA.
Under the same conditions, the semantic segmentation head accounts for 0.14\%, 0.11\%, 0.04\%, and 0.25\% of the full forward latency on \SemanticKITTI{}, \Waymo{}, \nuScenes{}, and \ScanNet{}, respectively.
Parameter counts cover the entire model, including the encoder, decoder, and prediction head.
For the semantic segmentation comparison tables, we report the pooled (\(s>1\)) parameter count for each \methodname variant; the count differs slightly from the unpooled (\(s=1\)) case due to the pooling and return projections.
Peak allocated GPU memory in \cref{fig:global-pooling-stride}(c) is measured in a separate forward pass under the same conditions using \texttt{torch.cuda.max\_memory\_allocated()}.
Forward hooks delimit non-overlapping regions; we reset the peak-memory counter at each region's entry and read it on exit, so the reported peaks include memory already allocated at region entry.

\subsection{Instance Segmentation}
\label{sec:suppl-instance-segmentation}

\paragraph{Datasets and Training Settings.}
Instance segmentation uses the PointGroup~\citep{Jiang_2020_CVPR} framework on \ScanNet{}~\citep{Dai_2017_CVPR} and \ScanNetTwoHundred{}~\citep{10.1007/978-3-031-19827-4_8}.
The training settings are summarized alongside those for indoor semantic segmentation in \cref{tab:indoor-training-settings}.

\paragraph{Model Adaptation.}
The backbone includes both the encoder and the feature decoder that restores spatial resolution before the PointGroup head.
PTv3~\citep{Wu_2024_CVPR} retains its hierarchical decoder with attention blocks.
The asterisk (*) denotes variants augmented with additional attention and convolution blocks in LitePT-S*~\citep{Yue_2026_CVPR} and \methodSstar{}.
LitePT-S* uses attention at its coarsest decoder stage and convolutions at the three finer stages.
\methodSstar{} appends two global attention blocks to the encoder and uses two convolutional blocks at each of its three decoder stages without decoder attention.
Volt~\citep{yilmaz2026volt} restores voxel features through a lightweight decoder based on linear projection and voxel shuffling.
Its published instance segmentation model additionally uses an SPFormer~\citep{Sun_2023} task decoder.
For our reproduction, we retain the voxel feature decoder and replace SPFormer with the common PointGroup head.
For PTv3, LitePT-S*, and \methodSstar{}, the backbone and PointGroup head are jointly trained from scratch for instance segmentation.
In contrast, following Volt's pretraining protocol, we initialize its backbone and feature decoder from the EMA weights of its \ScanNet{} or \ScanNetTwoHundred{} semantic segmentation model trained with CNN distillation, using the checkpoint corresponding to each dataset.
The pretrained backbone and feature decoder are then jointly fine-tuned with the PointGroup head without further EMA updates.

\paragraph{Evaluation.}
We follow the instance segmentation evaluation protocols of PTv3~\citep{Wu_2024_CVPR} and LitePT~\citep{Yue_2026_CVPR}, reporting mAP$_{25}$, mAP$_{50}$, and mAP on \ScanNet{} and \ScanNetTwoHundred{}.
All results in the main comparison are evaluated without EMA.

\subsection{Test-Time Augmentation and Fragments}
\label{sec:suppl-tta-fragments}

\begin{figure}[H]
  \centering
  \setlength{\abovecaptionskip}{3pt}
  \includegraphics[width=0.320\linewidth]{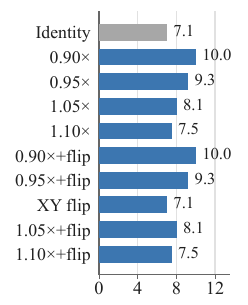}\hfill
  \includegraphics[width=0.320\linewidth]{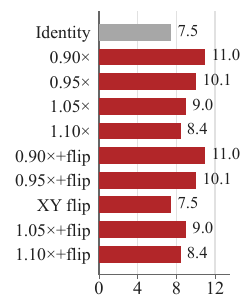}\hfill
  \includegraphics[width=0.320\linewidth]{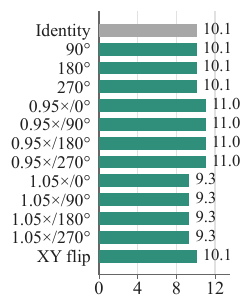}\par
  {\normalsize\strut Mean fragments / view\par}
  \vspace{3pt}
  \makebox[0.320\linewidth][c]{\normalsize\strut (a) \SemanticKITTI{}}\hfill
  \makebox[0.320\linewidth][c]{\normalsize\strut (b) \nuScenes{}}\hfill
  \makebox[0.320\linewidth][c]{\normalsize\strut (c) \ScanNet{}}\par
  \caption{Fragment counts across TTA views for \SemanticKITTI{}, \nuScenes{}, and \ScanNet{}.
Each fragment requires one model forward at batch size one.}
  \label{fig:tta-fragments}
\end{figure}

As described in Appendix~\ref{sec:suppl-semantic-segmentation}, a TTA view is a transformed input scene, whereas fragments are test-time voxelized realizations obtained by selecting different representative points from occupied voxels rather than by spatially partitioning the scene.
At batch size one, each fragment requires a separate forward, with predictions aggregated first across fragments and then across TTA views.
\cref{fig:tta-fragments} shows the mean number of fragments per TTA view; across all TTA views, the total number of fragment forwards is 11.9$\times$, 12.3$\times$, and 13.1$\times$ that of the identity view on \SemanticKITTI{}, \nuScenes{}, and \ScanNet{}, respectively.
Because of this additional cost, our main comparisons use results without TTA.
Disabling TTA removes only the transformed views; fragment aggregation is retained to reduce sensitivity to the representative points selected during test-time voxelization.

\section{Additional Results and Ablations}
\label{sec:suppl-ablations}

\begin{table}[H]
  \caption{Semantic segmentation w/ TTA and w/o TTA (\mIoU / \mAcc{}, \%).
  Bold and underline indicate the best and second-best results, respectively, among the baselines and \methodS{} within each TTA setting.}
  \label{tab:semantic-segmentation-full}
  \centering
  \small
  \newcommand{\faded}[1]{\textcolor{black!55}{#1}}
  \setlength{\tabcolsep}{4.1pt}
  \begin{tabular}{@{}c@{\hspace{2.5pt}}c@{\hspace{4pt}}lrrrrrrrrr@{}}
    \toprule
    & & & & \multicolumn{2}{c}{\SemKITTI{}} & \multicolumn{2}{c}{\Waymo} & \multicolumn{2}{c}{\nuScenes} & \multicolumn{2}{c}{\ScanNet} \\
    \cmidrule(lr){5-6}\cmidrule(lr){7-8}\cmidrule(lr){9-10}\cmidrule(l){11-12}
    & & Method & \#Params & \mIoU & \mAcc{} & \mIoU & \mAcc{} & \mIoU & \mAcc{} & \mIoU & \mAcc{} \\
    \midrule
    \multirow{6}{*}{\rotatebox[origin=c]{90}{\textit{w/ TTA}}}
    & \multirow{6}{*}{{\color{black!30}\rule{0.8pt}{12ex}}}
    & \faded{PTv3~\citep{Wu_2024_CVPR}} & \faded{46.1M} & \faded{71.0} & \faded{\underline{77.9}} & \faded{71.3} & \faded{80.5} & \faded{81.2} & \faded{87.5} & \faded{\textbf{77.6}} & \faded{\textbf{85.9}} \\
    & & \faded{LitePT-S~\citep{Yue_2026_CVPR}} & \faded{12.7M} & \faded{\textbf{71.3}} & \faded{77.7} & \faded{\textbf{73.1}} & \faded{\textbf{83.8}} & \faded{\textbf{82.2}} & \faded{\textbf{88.1}} & \faded{76.5} & \faded{84.2} \\
    & & \faded{Volt-S~\citep{yilmaz2026volt}} & \faded{23.7M} & \faded{70.5} & \faded{77.6} & \faded{71.5} & \faded{80.9} & \faded{81.0} & \faded{87.7} & \faded{\underline{77.3}} & \faded{\underline{85.5}} \\
    & & \faded{\textbf{\methodS{} (Ours)}} & \faded{14.1M} & \faded{\underline{71.2}} & \faded{\textbf{78.1}} & \faded{\underline{72.9}} & \faded{\underline{83.4}} & \faded{\underline{81.8}} & \faded{\underline{87.9}} & \faded{76.9} & \faded{84.9} \\
    \cmidrule(l){3-12}
    & & \faded{\textbf{\methodB{} (Ours)}} & \faded{51.8M} & \faded{71.8} & \faded{78.1} & \faded{73.0} & \faded{84.5} & \faded{81.7} & \faded{87.8} & \faded{77.4} & \faded{85.4} \\
    & & \faded{\textbf{\methodL{} (Ours)}} & \faded{89.4M} & \faded{71.6} & \faded{78.0} & \faded{73.1} & \faded{82.9} & \faded{82.4} & \faded{88.3} & \faded{77.7} & \faded{85.7} \\
    \midrule
    \multirow{6}{*}{\rotatebox[origin=c]{90}{\textit{w/o TTA}}}
    & \multirow{6}{*}{{\color{black!30}\rule{0.8pt}{12ex}}}
    & PTv3~\citep{Wu_2024_CVPR} & 46.1M & 69.6 & \underline{77.2} & 70.7 & 81.3 & 79.6 & 86.6 & \textbf{76.5} & \textbf{85.3} \\
    & & LitePT-S~\citep{Yue_2026_CVPR} & 12.7M & \underline{69.8} & 76.8 & \underline{71.7} & \underline{82.8} & 80.3 & 87.1 & 75.5 & 83.4 \\
    & & Volt-S~\citep{yilmaz2026volt} & 23.7M & 69.4 & 77.0 & 70.5 & 80.3 & \underline{80.5} & \textbf{87.5} & \underline{76.4} & \underline{84.9} \\
    & & \textbf{\methodS{} (Ours)} & 14.1M & \textbf{70.0} & \textbf{77.3} & \textbf{72.1} & \textbf{83.0} & \textbf{80.6} & \underline{87.4} & 76.0 & 84.3 \\
    \cmidrule(l){3-12}
    & & \textbf{\methodB{} (Ours)} & 51.8M & 70.4 & 77.2 & 72.1 & 84.1 & 81.1 & 87.3 & 76.4 & 84.9 \\
    & & \textbf{\methodL{} (Ours)} & 89.4M & 70.6 & 77.4 & 72.4 & 83.7 & 81.5 & 87.9 & 76.7 & 85.1 \\
    \bottomrule
  \end{tabular}
\end{table}

\paragraph{Semantic Segmentation with TTA.}
\cref{tab:semantic-segmentation-full} reports the complete semantic segmentation results with and without TTA.
TTA improves \methodS{} by 0.8--1.2 \mIoU{} points across the four datasets, while our main comparison focuses on results without TTA due to the higher evaluation cost discussed in Appendix~\ref{sec:suppl-tta-fragments}.

\begin{table}[H]
  \caption{Serialization order ablation of \methodS{} (w/o TTA, \mIoU / \mAcc{}, \%).}
  \label{tab:serialization-order-ablation}
  \centering
  \small
  \setlength{\tabcolsep}{3.5pt}
  \begin{tabular}{@{}lrrrrrrrrr@{}}
    \toprule
     & \multicolumn{3}{c}{\SemKITTI{}} & \multicolumn{3}{c}{\nuScenes} & \multicolumn{3}{c}{\ScanNet} \\
    \cmidrule(lr){2-4}\cmidrule(lr){5-7}\cmidrule(l){8-10}
    Order & Latency (ms) & \mIoU & \mAcc{} & Latency (ms) & \mIoU & \mAcc{} & Latency (ms) & \mIoU & \mAcc{} \\
    \midrule
    \texttt{Z} & 27.83 & 69.2 & 76.3 & 18.92 & 80.4 & 87.0 & 17.51 & 75.4 & 83.9 \\
    \rowcolor{black!10}
    \texttt{TZ} & 28.20 & 70.0 & 77.3 & 19.16 & 80.6 & 87.4 & 17.48 & 76.0 & 84.3 \\
    \texttt{H} & 30.15 & 69.1 & 76.8 & 21.34 & 80.4 & 87.0 & 18.97 & 75.5 & 83.8 \\
    \texttt{TH} & 30.22 & 69.2 & 76.1 & 21.02 & 80.3 & 87.2 & 18.95 & 75.2 & 84.2 \\
    \bottomrule
  \end{tabular}
\end{table}

\paragraph{Serialization Order.}
\label{sec:suppl-serialization-order}
We use Trans Z-order (\texttt{TZ}) as the default serialization pattern.
Across all three datasets, performance varies only modestly across the tested fixed orders, with \texttt{TZ} achieving the highest \mIoU{}.
Serialization overhead consists of two main components: code construction and stage layout.
Code construction encodes coordinates into serialization codes, sorts them to obtain the serialization order, and builds the inverse mapping, while the stage layout reorders features and coordinates once when entering the attention stage and restores the original order when leaving it.
The stage layout cost remains similar across serialization orders, whereas code construction is more expensive for the Hilbert-based alternatives in our implementation.

\begin{table}[H]
\caption{Median axis ranges over the complete validation sets.
Scene ranges are measured in metres, while window and global ranges are measured in cells of their respective attention grids.}
  \label{tab:serialized-window-geometry}
  \centering
  \small
  \setlength{\tabcolsep}{5.5pt}
  \begin{tabular}{@{}lrrrrrrrrr@{}}
    \toprule
    & \multicolumn{3}{c}{Scene [m]} & \multicolumn{3}{c}{Window extent [cells]} & \multicolumn{3}{c}{Global extent [cells]} \\
    \cmidrule(lr){2-4}\cmidrule(lr){5-7}\cmidrule(l){8-10}
    Dataset & $x$ & $y$ & $z$ & $x$ & $y$ & $z$ & $x$ & $y$ & $z$ \\
    \midrule
    \SemKITTI{} & 157.3 & 109.8 & 9.7 & 79 & 63 & 11 & 99 & 69 & 7 \\
    \Waymo & 148.6 & 132.4 & 8.5 & 72 & 63 & 14 & 93 & 83 & 6 \\
    \nuScenes & 145.0 & 188.3 & 24.3 & 126 & 120 & 31 & 182 & 236 & 31 \\
    \ScanNet & 5.6 & 5.1 & 2.5 & 30 & 23 & 16 & 35 & 32 & 16 \\
    \bottomrule
  \end{tabular}
\end{table}

\paragraph{Scene and Attention Geometry.}
\label{sec:suppl-attention-geometry}
For window attention, we measure the axis ranges of each full 1,024-token serialized window on the attention-stage grid.
Across all four datasets, these windows span wider ranges along the horizontal axes than along the vertical axis.
For global attention, we measure the axis ranges of the scene-wide token set after dataset-specific pooling when \(s>1\), or at the original attention-stage resolution when \(s=1\).
A similar horizontal-vertical imbalance is observed on the resulting global-attention grid.
Because window and global ranges are measured on grids with different cell sizes, their cell counts should not be compared directly as physical distances.

\section{Limitations}
\label{sec:suppl-limitations}

This study focuses on establishing the effectiveness of interleaving window and global attention rather than exhaustively optimizing their arrangement.
Although we evaluate several block layouts, the best schedule may depend on the dataset, task, or model scale, and a more systematic search for when global interaction should be introduced remains an open direction.

Global attention is currently made more efficient through spatial grid pooling, which provides a simple way to compress the token set before scene-wide interaction.
More generally, the key challenge is to reduce the number of tokens while preserving the information needed for effective scene-wide interaction.
Exploring alternative pooling, token selection, or learned compact representations may further improve this efficiency-accuracy trade-off.

\clearpage
\section{Qualitative Results}
\label{sec:suppl-qualitative}

\begin{figure}[H]
  \centering
  \begingroup
  \setlength{\tabcolsep}{0pt}
  \newlength{\semkittipanelwidth}
  \setlength{\semkittipanelwidth}{0.313\linewidth}
  \newlength{\semkittipanelheight}
  \newcommand{\semkittiscenelabel}[1]{%
    \begingroup
    \setlength{\unitlength}{\semkittipanelheight}%
    \begin{picture}(0.15,1)
      \put(0.075,0.5){\makebox(0,0){%
        \rotatebox{90}{\fontsize{7}{8}\selectfont\texttt{08\_#1}}}}%
    \end{picture}%
    \endgroup}
  \newcommand{\semkittirow}[4]{%
    \settoheight{\semkittipanelheight}{#2}%
    \semkittiscenelabel{#1} &
    #2 & #3 & #4}
  {\fontsize{7.4}{10}\selectfont
    \input{pics/semkitti_qualitative/legend}\par}
  \vspace{6pt}
  \begin{tabular*}{\linewidth}{@{\extracolsep{\fill}}cccc@{}}
    \semkittirow{000900}
      {\includegraphics[width=\semkittipanelwidth]{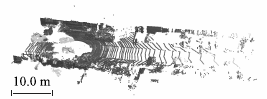}}
      {\includegraphics[width=\semkittipanelwidth]{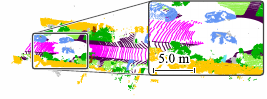}}
      {\includegraphics[width=\semkittipanelwidth]{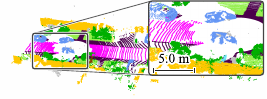}} \\[5.5pt]
    \semkittirow{001500}
      {\includegraphics[width=\semkittipanelwidth]{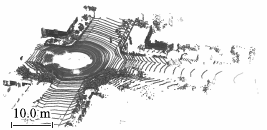}}
      {\includegraphics[width=\semkittipanelwidth]{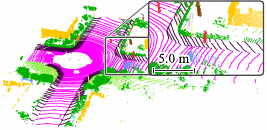}}
      {\includegraphics[width=\semkittipanelwidth]{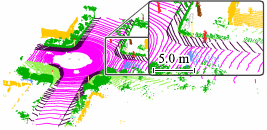}} \\[5.5pt]
    \semkittirow{001700}
      {\includegraphics[width=\semkittipanelwidth]{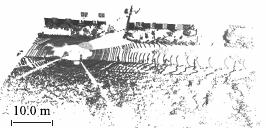}}
      {\includegraphics[width=\semkittipanelwidth]{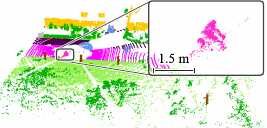}}
      {\includegraphics[width=\semkittipanelwidth]{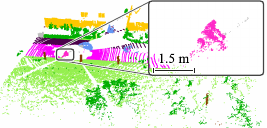}} \\[5.5pt]
    \semkittirow{002700}
      {\includegraphics[width=\semkittipanelwidth]{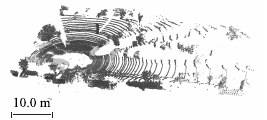}}
      {\includegraphics[width=\semkittipanelwidth]{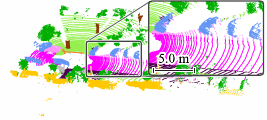}}
      {\includegraphics[width=\semkittipanelwidth]{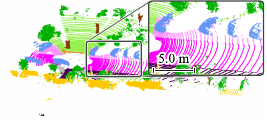}} \\[5.5pt]
    \semkittirow{003400}
      {\includegraphics[width=\semkittipanelwidth]{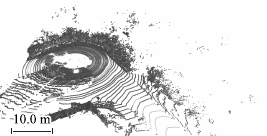}}
      {\includegraphics[width=\semkittipanelwidth]{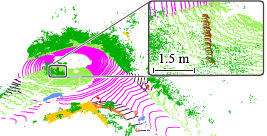}}
      {\includegraphics[width=\semkittipanelwidth]{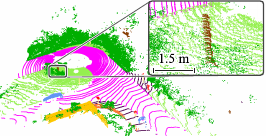}} \\[5.5pt]
    \semkittirow{004000}
      {\includegraphics[width=\semkittipanelwidth]{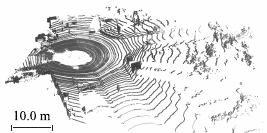}}
      {\includegraphics[width=\semkittipanelwidth]{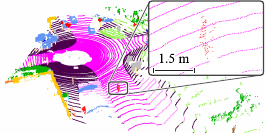}}
      {\includegraphics[width=\semkittipanelwidth]{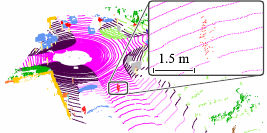}} \\[5.5pt]
    \semkittirow{002400}
      {\includegraphics[width=\semkittipanelwidth]{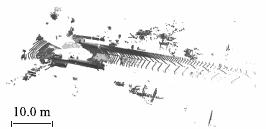}}
      {\includegraphics[width=\semkittipanelwidth]{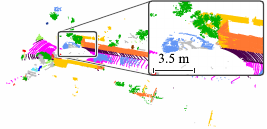}}
      {\includegraphics[width=\semkittipanelwidth]{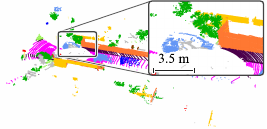}} \\[4pt]
    & (a) Input & (b) Ours & (c) GT
  \end{tabular*}
  \endgroup
  \caption{Qualitative semantic segmentation on \SemanticKITTI{}.
Columns show (a) input point clouds, (b) \methodS{} predictions, and (c) ground truth.
Insets highlight representative regions.}
\label{fig:semkitti-qualitative}
\end{figure}

\clearpage
\begin{figure}[H]
  \centering
  \begingroup
  \setlength{\tabcolsep}{0pt}
  \newlength{\scannetpanelwidth}
  \setlength{\scannetpanelwidth}{0.313\linewidth}
  \newlength{\scannetpanelheight}
  \setlength{\scannetpanelheight}{0.173889\linewidth}
  \newcommand{\scannetscenelabel}[1]{%
    \begingroup
    \setlength{\unitlength}{\scannetpanelheight}%
    \begin{picture}(0.15,1)
      \put(0.075,0.5){\makebox(0,0){%
        \rotatebox{90}{\fontsize{7}{8}\selectfont\texttt{\detokenize{#1}}}}}%
    \end{picture}%
    \endgroup}
  \newcommand{\scannetrow}[4]{%
    \scannetscenelabel{#1} &
    #2 & #3 & #4}
  {\fontsize{7.4}{10}\selectfont
    \input{pics/scannet_qualitative/legend}\par}
  \vspace{6pt}
  \begin{tabular*}{\linewidth}{@{\extracolsep{\fill}}cccc@{}}
    \scannetrow{scene0559_02}
      {\includegraphics[width=\scannetpanelwidth]{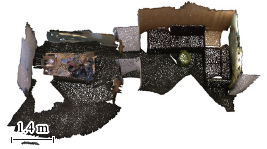}}
      {\includegraphics[width=\scannetpanelwidth]{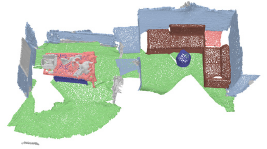}}
      {\includegraphics[width=\scannetpanelwidth]{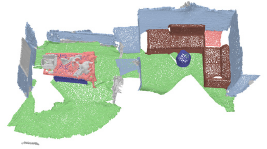}} \\[6pt]
    \scannetrow{scene0278_00}
      {\includegraphics[width=\scannetpanelwidth]{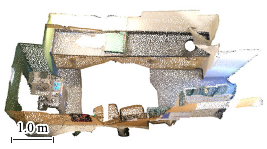}}
      {\includegraphics[width=\scannetpanelwidth]{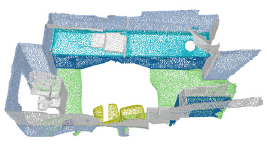}}
      {\includegraphics[width=\scannetpanelwidth]{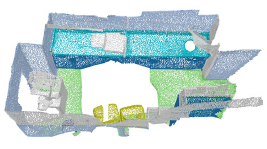}} \\[6pt]
    \scannetrow{scene0435_03}
      {\includegraphics[width=\scannetpanelwidth]{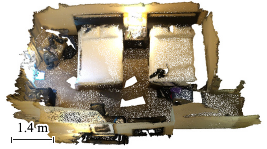}}
      {\includegraphics[width=\scannetpanelwidth]{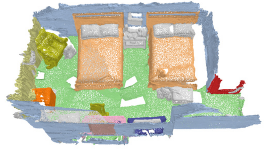}}
      {\includegraphics[width=\scannetpanelwidth]{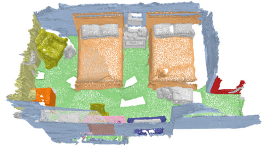}} \\[6pt]
    \scannetrow{scene0077_01}
      {\includegraphics[width=\scannetpanelwidth]{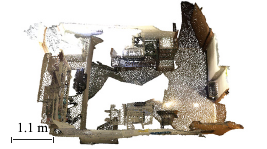}}
      {\includegraphics[width=\scannetpanelwidth]{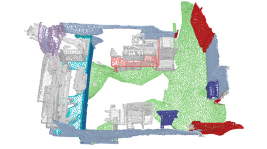}}
      {\includegraphics[width=\scannetpanelwidth]{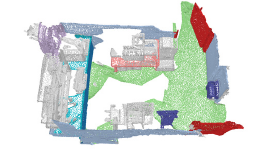}} \\[6pt]
    \scannetrow{scene0378_00}
      {\includegraphics[width=\scannetpanelwidth]{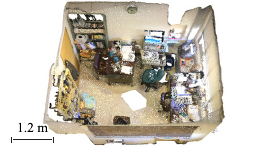}}
      {\includegraphics[width=\scannetpanelwidth]{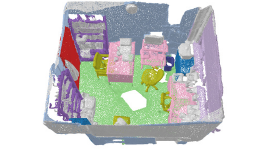}}
      {\includegraphics[width=\scannetpanelwidth]{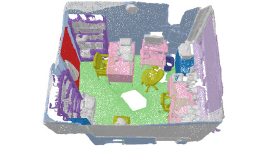}} \\[6pt]
    \scannetrow{scene0474_05}
      {\includegraphics[width=\scannetpanelwidth]{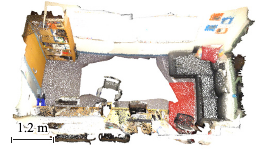}}
      {\includegraphics[width=\scannetpanelwidth]{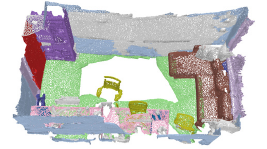}}
      {\includegraphics[width=\scannetpanelwidth]{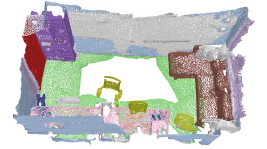}} \\[6pt]
    \scannetrow{scene0095_01}
      {\includegraphics[width=\scannetpanelwidth]{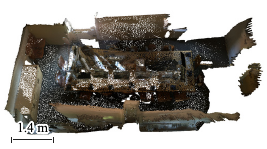}}
      {\includegraphics[width=\scannetpanelwidth]{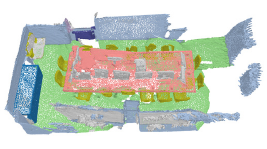}}
      {\includegraphics[width=\scannetpanelwidth]{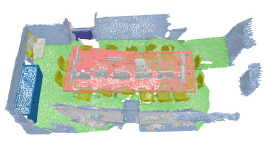}} \\[4pt]
    & (a) Input & (b) Ours & (c) GT
  \end{tabular*}
  \endgroup
  \caption{Qualitative semantic segmentation on \ScanNet{}.
Columns show (a) input point clouds, (b) \methodS{} predictions, and (c) ground truth.}
\label{fig:scannet-qualitative}
\end{figure}

\clearpage
\begin{figure}[H]
  \centering
  \begingroup
  \setlength{\tabcolsep}{0pt}
  \newlength{\scannetinstancepanelwidth}
  \setlength{\scannetinstancepanelwidth}{0.313\linewidth}
  \newlength{\scannetinstancepanelheight}
  \setlength{\scannetinstancepanelheight}{0.173889\linewidth}
  \newcommand{\scannetinstancescenelabel}[1]{%
    \begingroup
    \setlength{\unitlength}{\scannetinstancepanelheight}%
    \begin{picture}(0.15,1)
      \put(0.075,0.5){\makebox(0,0){%
        \rotatebox{90}{\fontsize{7}{8}\selectfont\texttt{\detokenize{#1}}}}}%
    \end{picture}%
    \endgroup}
  \newcommand{\scannetinstancerow}[4]{%
    \scannetinstancescenelabel{#1} &
    #2 & #3 & #4}
  \begin{tabular*}{\linewidth}{@{\extracolsep{\fill}}cccc@{}}
    \scannetinstancerow{scene0427_00}
      {\includegraphics[width=\scannetinstancepanelwidth]{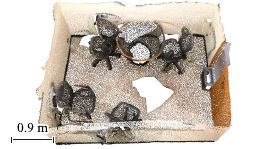}}
      {\includegraphics[width=\scannetinstancepanelwidth]{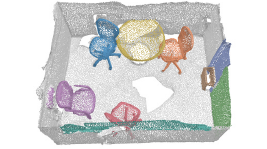}}
      {\includegraphics[width=\scannetinstancepanelwidth]{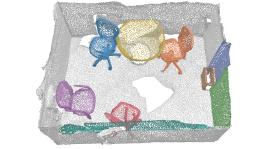}} \\[6pt]
    \scannetinstancerow{scene0621_00}
      {\includegraphics[width=\scannetinstancepanelwidth]{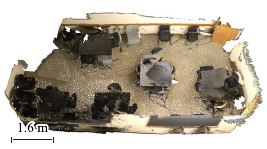}}
      {\includegraphics[width=\scannetinstancepanelwidth]{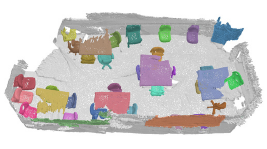}}
      {\includegraphics[width=\scannetinstancepanelwidth]{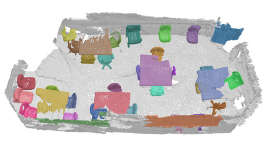}} \\[6pt]
    \scannetinstancerow{scene0488_01}
      {\includegraphics[width=\scannetinstancepanelwidth]{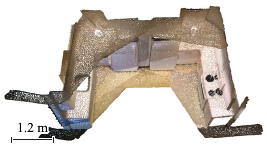}}
      {\includegraphics[width=\scannetinstancepanelwidth]{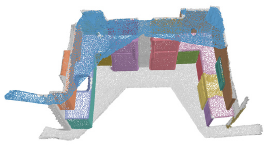}}
      {\includegraphics[width=\scannetinstancepanelwidth]{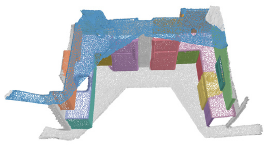}} \\[6pt]
    \scannetinstancerow{scene0599_00}
      {\includegraphics[width=\scannetinstancepanelwidth]{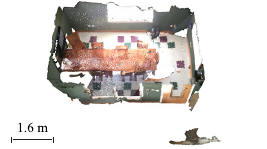}}
      {\includegraphics[width=\scannetinstancepanelwidth]{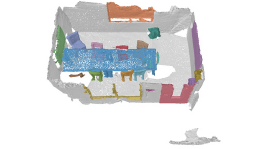}}
      {\includegraphics[width=\scannetinstancepanelwidth]{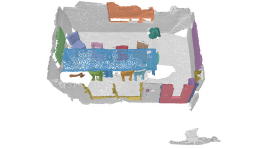}} \\[6pt]
    \scannetinstancerow{scene0217_00}
      {\includegraphics[width=\scannetinstancepanelwidth]{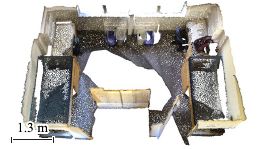}}
      {\includegraphics[width=\scannetinstancepanelwidth]{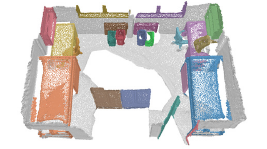}}
      {\includegraphics[width=\scannetinstancepanelwidth]{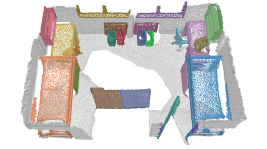}} \\[6pt]
    \scannetinstancerow{scene0578_02}
      {\includegraphics[width=\scannetinstancepanelwidth]{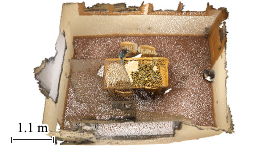}}
      {\includegraphics[width=\scannetinstancepanelwidth]{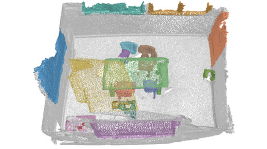}}
      {\includegraphics[width=\scannetinstancepanelwidth]{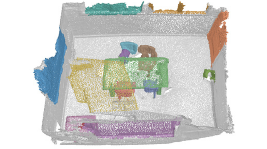}} \\[6pt]
    \scannetinstancerow{scene0645_02}
      {\includegraphics[width=\scannetinstancepanelwidth]{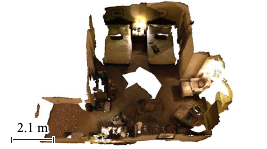}}
      {\includegraphics[width=\scannetinstancepanelwidth]{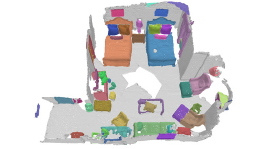}}
      {\includegraphics[width=\scannetinstancepanelwidth]{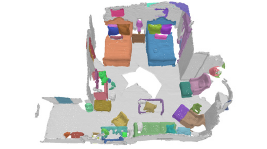}} \\[4pt]
    & (a) Input & (b) Ours & (c) GT
  \end{tabular*}
  \endgroup
  \caption{Qualitative instance segmentation on \ScanNetTwoHundred{}.
Columns show (a) input point clouds, (b) \methodSstar{} predictions, and (c) ground truth.
Colors distinguish instances rather than semantic classes.
For visualization, prediction-GT pairs matched one-to-one within the same class at mask IoU $>0.5$ share colors; unmatched instances receive distinct colors.
Gray indicates background or unassigned points.}
\label{fig:scannet200-instance-qualitative}
\end{figure}

\end{document}